\documentclass[10pt,twocolumn,letterpaper]{article}

\usepackage[pagenumbers]{cvpr} 

\newcommand{\red}[1]{{\color{red}#1}}

\definecolor{cvprblue}{rgb}{0.21,0.49,0.74}

\usepackage[normalem]{ulem}
\usepackage{multirow}
\usepackage{makecell}
\usepackage{duckuments}
\usepackage{placeins}
\usepackage{amssymb}
\usepackage{pifont}
\usepackage[dvipsnames]{xcolor}

\newcommand{\greencheckmark}[1]{\textcolor{ForestGreen}{\checkmark}}
\newcommand{\redcrossmark}[1]{\textcolor{red}{\crossmark}}

\usepackage{mdframed}
\definecolor{json_bg}{RGB}{206, 242, 245}
\definecolor{yaml_bg}{RGB}{206, 245, 230}
\definecolor{prompt_bg}{RGB}{234, 234, 251}

\definecolor{cvprblue}{rgb}{0.21,0.49,0.74}
\definecolor{laucolor}{RGB}{0, 102, 204}

\usepackage{fancyvrb}

\usepackage{listings}
\usepackage{./fancyhdr}

\usepackage[toc,page,header]{appendix}
\usepackage{minitoc}

\usepackage[hang]{footmisc}  
\usepackage[pagebackref,breaklinks,colorlinks,allcolors=cvprblue]{hyperref}

\def\paperID{5712} 
\def\confName{CVPR}
\def\confYear{2025}

\title{MicroVQA: A Multimodal Reasoning Benchmark for Microscopy-Based Scientific Research}

\author{
James Burgess*$^{1}$ \quad
Jeffrey J Nirschl*$^{1}$ \quad
Laura Bravo-S\'anchez$^{1,*}$ \\ 
Alejandro Lozano$^{1}$\quad
Sanket Rajan Gupte$^{1}$\quad
Jesus G. Galaz-Montoya$^{1}$\quad
Yuhui Zhang$^{1}$ \\ 
Yuchang Su$^{2}$\quad
Disha Bhowmik$^{3}$\quad 
Zachary Coman$^{3}$\quad
Sarina M. Hasan$^{4}$ \\
Alexandra Johannesson$^{5}$ \quad
William D. Leineweber$^{1}$ \quad
Malvika G Nair$^{3}$ \quad 
Ridhi Yarlagadda$^{3}$\\
Connor Zuraski$^{1}$\quad 
Wah Chiu$^{1}$\quad
Sarah Cohen$^{3}$\quad
Jan N. Hansen$^{1}$\quad
Manuel D Leonetti$^{6}$ \\
Chad Liu$^{6}$\quad
Emma Lundberg$^{1,5,6}$\quad
Serena Yeung-Levy$^{1,6}$
}

\begin{document}

\twocolumn[{
    \renewcommand\twocolumn[1][]{#1}
    \maketitle
    \doparttoc 
    \faketableofcontents
    
    \vspace{-3em}
    \iftoggle{cvprfinal}{
        \begin{center}
            $^{1}$Stanford University, $^{2}$Tsinghua University, $^{3}$University of North Carolina at Chapel Hill, \\
            $^{4}$Princeton University, $^{5}$KTH Royal Institute of Technology, $^{6}$Chan Zuckerberg Biohub Network \\
        \end{center}
    }
    \centering
    \includegraphics[width=\textwidth,clip,trim=0cm 0cm 0cm 0cm]{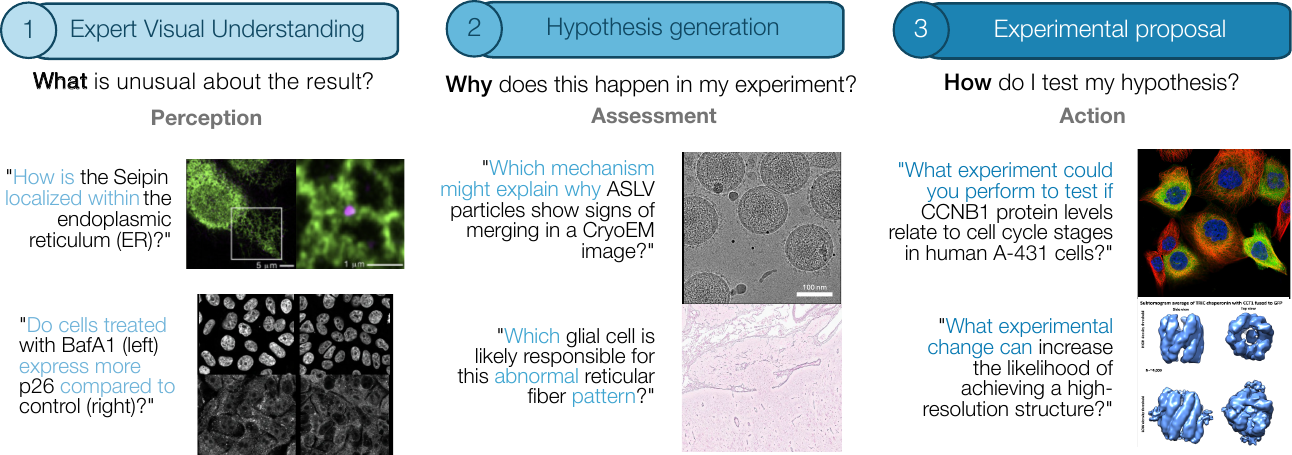}
    \captionof{figure}{A scientific experimentation workflow drives discovery: researchers analyze experiments, develop hypotheses, and design further experiments to test their ideas. We release MicroVQA, a visual question answering (VQA) benchmark to test these three tasks in the context of biological microscopy.  Each of the 1,042 samples is created by a biology expert, and transformed into a multiple choice question (MCQ).
    }
    \label{fig:pull}
    \vspace{1em}
    }]

\renewcommand{\thefootnote}{}  
\footnotetext{Correspondence \texttt{\{jmhb,jnirschl,lmbravo\}@stanford.edu} \\$^{*}$These authors contributed equally}

\begin{abstract}
Scientific research demands sophisticated reasoning over multimodal data, a challenge especially prevalent in biology. Despite recent advances in multimodal large language models (MLLMs) for AI-assisted research, existing multimodal reasoning benchmarks only target up to college-level difficulty, while research-level benchmarks emphasize lower-level perception, falling short of the complex multimodal reasoning needed for scientific discovery. To bridge this gap, we introduce MicroVQA, a visual-question answering (VQA) benchmark designed to assess three reasoning capabilities vital in research workflows: expert image understanding, hypothesis generation, and experiment proposal. MicroVQA consists of 1,042 multiple-choice questions (MCQs) curated by biology experts across diverse microscopy modalities, ensuring VQA samples represent real scientific practice. In constructing the benchmark, we find that standard MCQ generation methods induce language shortcuts, motivating a new two-stage pipeline: an optimized LLM prompt structures question-answer pairs into MCQs; then, an agent-based `RefineBot' updates them to remove shortcuts. Benchmarking on state-of-the-art MLLMs reveal a peak performance of 53\%; models with smaller LLMs only slightly underperform top models, suggesting that language-based reasoning is less challenging than multimodal reasoning; and tuning with scientific articles enhances performance. Expert analysis of chain-of-thought responses shows that perception errors are the most frequent, followed by knowledge errors and then overgeneralization errors. These insights highlight the challenges in multimodal scientific reasoning, showing MicroVQA is a valuable resource advancing AI-driven biomedical research. MicroVQA is available  \href{https://huggingface.co/datasets/jmhb/microvqa}{here}, project \href{https://jmhb0.github.io/microvqa}{here}.
\end{abstract}

\vspace{-1em}
\section{Introduction}
\label{sec:intro}
The transformative potential of AI in scientific discovery lies in its ability to interpret and reason over complex, multimodal data while integrating specialist domain knowledge. Recently, multimodal large language models (MLLMs) have demonstrated notable capabilities in data comprehension, recall, and multi-step reasoning across various modalities on exam-style questions \cite{alayrac2022flamingo, achiam2023gpt4, team2023gemini, anthropic2024claude3, liu2024visual, wang2024exploring}. This progress has generated enthusiasm for scientific AI applications, ranging from multimodal chat AI assistants \cite{hope2023computational, thapa2023chatgpt, zheng2023chatgpt, xie2023large} to AI agents capable of orchestrating complex experiments \cite{gao2024empowering, boiko2023autonomous, m2024augmenting, carpenter2023smart}. However, while MLLMs perform well on general exam tasks, they often lack the specialized, contextual reasoning needed for real scientific challenges, where tasks require advanced image-based reasoning, analysis, and hypothesis-driven experimentation.

In recent years, efforts to apply scientific MLLMs across a range of domains have laid an essential groundwork. Advances have been achieved in image understanding \cite{lozano2024mu, li2024mmsci, laurent2024lab, li2024chemvlm, maruf2024vlm4bio}, literature analysis \cite{skarlinski2024language, wadden2024sciriff, li2024mmsci, laurent2024lab}, and problem-solving at school and early college levels \cite{lu2022learn, yue2024mmmu}. In applied settings, AI has facilitated scientific tasks in biology  \cite{narayanan2024aviary, gottweis2025towards, swanson2024virtual, royer2024omega}, chemistry \cite{boiko2023autonomous, m2024augmenting, kang2024chatmof}, software \cite{tian2024scicode, royer2024omega}, and diagnostic pathology and radiology \cite{lu2024-pathchat}. However, there has been less progress on multimodal tasks requiring \textit{reasoning} in \textit{research-level} settings. The gap is due in part to a lack of multimodal benchmarks for scientific reasoning -- unlike college-level tasks where existing exams can form benchmarks, research tasks require expert curation. 

Microscopy provides a compelling use case for developing such a benchmark. It requires more than pattern recognition or factual recall -- it demands the ability to synthesize visual information with experimental context, formulate hypotheses, and propose experimental follow-ups. To evaluate these advanced capabilities, we introduce MicroVQA, a visual-question answering (VQA) benchmark tailored for multimodal scientific reasoning. Grounded in the real-world scientific practice of biological microscopy, MicroVQA incorporates 1,042 multiple-choice questions (MCQs) manually created by biological researchers across diverse disciplines, with each question taking more than 30 minutes to generate. Careful expert curation ensures the benchmark is relevant to real scientific applications.

MicroVQA is designed to evaluate three key reasoning tasks crucial to scientific inquiry: (1) \textit{expert image understanding} -- the ability to recognize scientifically relevant features and integrate contextual information about sample preparation; (2) \textit{hypothesis generation} -- the formulation of scientifically plausible explanations from experimental data; and (3) \textit{experiment proposal} -- the ability to suggest further experiments that could validate or refute hypotheses. These capabilities form a common workflow in scientific experimentation.
They demand both abductive reasoning -- inferring the best explanation from multiple possibilities -- and deductive reasoning -- moving from general principles to specific conclusions \cite{wang2024exploring}.

We tested standard methods to map open scientific VQA samples to multiple-choice questions (MCQs), but found that they induce language shortcuts -- the MCQs failed to truly test multimodal abilities. To ensure question quality and difficulty, we developed a two-stage pipeline for MCQ generation: first, an optimized LLM prompt structures QA pairs into well-formatted MCQs; then, a novel agent-based `RefineBot' system increases question difficulty by rewriting questions without language shortcuts. This ensures that MicroVQA's MCQs better test true scientific capabilities, rather than simple pattern recognition.

We benchmarked frontier MLLMs on MicroVQA, finding that the best performers achieved a modest peak of 53\%, which shows a significant gap between current models and expert-level scientific reasoning. The variation between different MLLMs was small; notably, smaller LLMs only slightly underperform their larger counterparts, suggesting that the key challenge for our MCQs is not language-only reasoning, but multimodal reasoning or knowledge. We also find that finetuning MLLMs on scientific publications improves MicroVQA performance. Finally, a group of biomedical researchers performed a detailed qualitative analysis into MLLM failure modes, uncovering three major failure modes: Expert perception is the most common, followed by knowledge errors, and then overgeneralization reasoning errors. This further supports that visual reasoning is a key challenge in MicroVQA. 

In summary, our contributions are as follows:
\begin{itemize}
    \item We introduce MicroVQA, a benchmark for multimodal reasoning in biological microscopy. We defined three key tasks for scientific research -- expert image understanding, hypothesis generation, and experiment proposal. Then, expert researchers curated 1,042 diverse VQA samples.
    \item We develop a novel two-stage MCQ generation pipeline to address the challenge of creating MCQs from long-format scientific questions without language shortcuts.
    \item We provide quantitative and qualitative benchmarking of frontier MLLMs, highlighting areas for improvement. 
\end{itemize}

\section{Related work}
\label{sec:related_work}
\paragraph{MLLM reasoning benchmarks} 
While \textit{perception} focuses on identifying an image's visual features, \textit{visual reasoning} often integrates prior knowledge to derive new conclusions using logical inference strategies like deduction, induction, abduction, and analogical reasoning \cite{wang2024exploring}.  Reasoning in MLLMs has focused on image-text VQA \cite{wang2024exploring, li2024survey, fu2023mme, yue2024mmmu, yue2024mmmu_pro, lu2022learn}. The most prominent examples in science are MMMU \cite{yue2024mmmu, yue2024mmmu_pro}, ScienceQA, \cite{lu2022learn} and MathVista \cite{lu2023mathvista}. While they do test reasoning, they are derived from exams up to the college level, while our benchmark emphasizes research-level difficulty. Other popular multimodal benchmarks test commonsense inference \cite{zellers2019recognition, thrush2022winoground}, spatial or relational awareness \cite{hudson2019gqa, zhang2019raven} and integrating prior knowledge \cite{marino2019ok}. Another line of work is visual abductive reasoning \cite{hessel2022abduction, bhagavatula2019abductive}, or finding the most likely explanation for an observation from incomplete information; this relates closely to hypothesis generation in this work. 

\paragraph{MLLMs in science}
Many works consider multimodal VQA in scientific or medical domains like microscopy images \cite{lozano2024mu}, biological images \cite{maruf2024vlm4bio}, medical images \cite{hu2024omnimedvqa, he2020pathvqa}, and chemical structure drawings \cite{li2024chemvlm}. Still, they mostly test more straightforward perception and recognition.  Figure comprehension for scientific articles does require more reasoning with multimodal data \cite{li2024mmsci, laurent2024lab, pramanick2024spiqa, li2024multimodal, li2023scigraphqa}; however, the images here are curated for publication, and the emphasis is on understanding content, rather than performing practical tasks. On the other hand, our benchmark includes realistic experiment images and tests more advanced reasoning like hypothesis generation and experiment planning. Scientific tasks have received much more attention in language-only LLMs \cite{rein2023gpqa, wang2023scibench, tian2024scicode,feng2024sciknoweval, jin2019pubmedqa, cai2024sciassess, laurent2024lab, royer2024omega, mirza2024large, sun2024scieval, skarlinski2024language, bianchi2025cardbiomedbench, zhao2025biomaze}, with GPQA being a notable work targeting PhD-level multi-step reasoning \cite{rein2023gpqa}, but these do not test multimodal reasoning, which is important to many scientific areas \cite{gao2024empowering}.  In terms of models, most MLLMs in the biomedical domain are targeted at clinical tasks \cite{li2024llava, zhang2024generalist, moor2023foundation, tu2024towards, saab2024capabilities, hu2024omnimedvqa}.

\paragraph{Multiple-choice question construction} 
A challenge in creating exams for education research and benchmarks in machine learning is generating multiple-choice questions with quality distractors \cite{gierl2017developing, alhazmi2024distractor}. Recent works have explored LLM-based generation methods, including zero-shot distractor prompting, few-shot prompting, LLM fine-tuning with high-quality examples, and prompting for known error types \cite{bitew2023distractor, feng2024exploring, fernandez2024divert, zhang2025automated}. However, a major challenge is that generated distractors are not sufficiently vision-centric -- MLLMs can often select the correct answer from the distractors without access to the image \cite{tong2024cambrian}. While this is partly explained by language biases on truly vision-centric MCQs \cite{goyal2017making, balepur2024artifacts}, others likely do fail to test visual understanding due to poor MCQ construction \citep{laurent2024lab} -- this informs our RefineBot MCQ generation method in \cref{sec:mcqgen-method}.

\section{The MicroVQA benchmark}
\label{sec:benchmark}

\begin{table}[t]
    \footnotesize
    \centering
    \caption{MicroVQA benchmark attributes.}    \label{tab:benchmark_attributes}
\begin{tabular}{lrl}
\toprule
\textbf{Dataset feature} & \multicolumn{1}{r}{\textbf{Value}} &  \\
\hline 
Total questions & 1,042 &  \\
Multi-image questions & 423 &  \\
\hline 
Avg. MCQ question length& 66&  \\
Avg. MCQ answer length& 15 &  \\
Avg. raw question length & 158 &  \\
Avg. raw answer length& 52 &  \\
\hline
Unique image sets & 255 &  \\
Image Modalities & Light, Fluoro, Electron &  \\
Image Scales & Tissue, Cell, Subcell, Atomic \\
\hline 
Organisms & 31 &  \\
Research areas & 33 & \\
\hline
Expert question creators & 12 &\\
Time to create 1 question & 30-40 mins &\\
Time to quality check 1 MCQ &	5 mins &\\
\bottomrule
\end{tabular}
\end{table}
\begin{figure}[t]
    \centering
    \includegraphics[width=\linewidth]{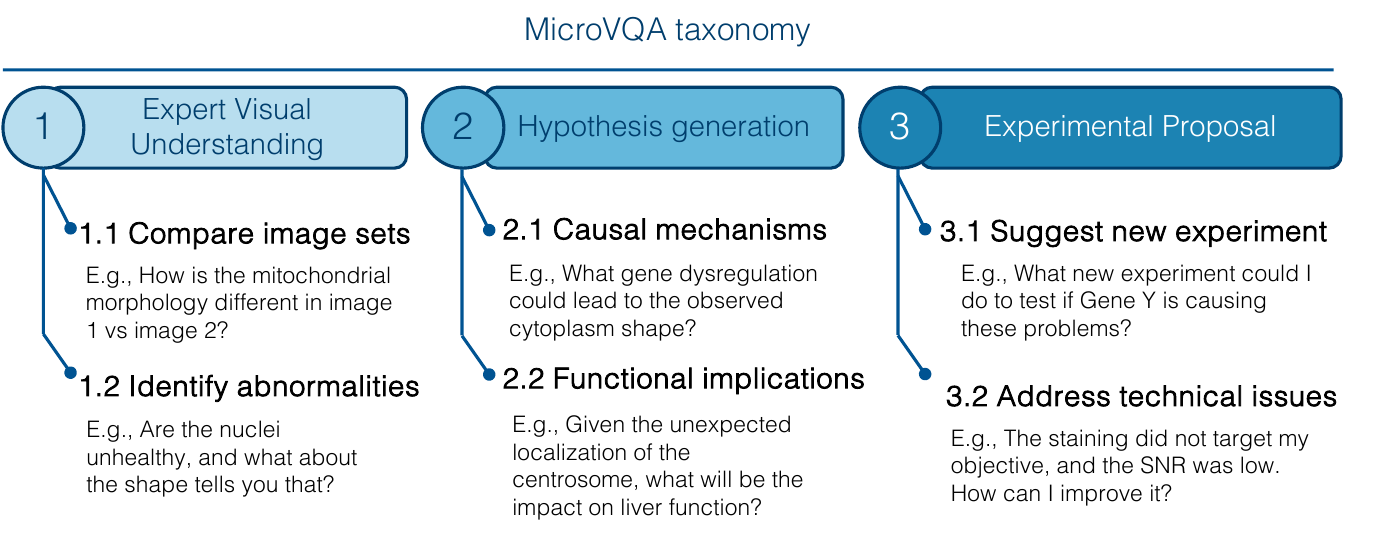}
    \vspace{-2em}
    \label{fig:taxonomy}
    \caption{MicroVQA taxonomy of sub-tasks.}
\end{figure}

Here, we present the MicroVQA benchmark for visual-question-answering (VQA) which advances multimodal scientific reasoning by tackling two major evaluation challenges. 
First, defining tasks is challenging because scientific research encompasses many complex activities \cite{hope2023computational} -- we therefore define three important tasks important to biomedical discovery. Second, VQA samples cannot be readily sourced from existing exams or classification datasets -- we therefore assembled a team of biologists to curate challenging reasoning problems. In \cref{sec:benchmark-overview}, we provide an overview of the MicroVQA benchmark; \cref{sec:benchmark-tasks} details the key tasks; and finally, \cref{sec:benchmark-reasoning} demonstrates that MicroVQA is at the forefront of measuring research-level reasoning in MLLMs. In the later \cref{sec:mcqgen}, we describe our approach for generating multiple-choice questions. 

\label{sec:benchmark-reasoning}

\subsection{Overview of MicroVQA}
\label{sec:benchmark-overview}
MicroVQA is a comprehensive dataset of 1,042 VQA triplets, manually curated by a team of expert biological researchers from diverse fields, and key attributes are summarized in \cref{tab:benchmark_attributes}. These expert-crafted questions test reasoning across three key tasks: expert understanding, hypothesis generation, and experiment proposal, and we additionally provide a taxonomy of sub-tasks in \cref{fig:taxonomy}. The questions cover a broad spectrum, ranging from biological issues, such as ``What is unusual about the cell shape?'' -- to technical imaging challenges like ``Is this feature due to low signal-to-noise ratio (SNR) or is it real?''. We provide the `raw' questions and answers written by experts, which are often long and nuanced, along with converted multiple-choice questions (MCQs) suitable for MLLM evaluation. 

The images encompass the most common microscopy modalities used in human biology: brightfield, fluorescence, and electron microscopy. The sample types span the full range of microscopic scales -- tissue,  cellular, subcellular, and atomic -- and emphasize organisms relevant to human biology and medically motivated tasks, namely human and mouse. Over 60\% of the samples have multiple images because comparison is essential in microscopy research, and because multichannel fluorescence images cannot be represented in standard RGB. All these attributes are provided as metadata tags to facilitate deeper error analysis, thereby enhancing MicroVQA's value as a resource for advancing AI capabilities in scientific research.

\subsection{MicroVQA scientific reasoning tasks}
\label{sec:benchmark-tasks}
To define specific reasoning tasks aimed at advancing biomedical research, we established the following criteria: (1) the tasks should use image-text inputs and text outputs suitable for MLLMs, (2) they should require higher-order reasoning beyond simple image processing, and (3) they should emphasize core experimental activities such as experimentation, analysis, and action \cite{hope2023computational}, rather than auxiliary tasks like literature review or writing. The tasks were developed through interviews with nine co-author PIs and postdocs specializing in microscopy research (\cref{sec:appendix-benchmark}). We identified three essential capabilities that we now introduce, with examples shown in \cref{fig:taxonomy}.

\paragraph{Expert visual understanding}
After performing an experiment, scientists must interpret data within its experimental context. In biological microscopy, this involves identifying patterns like protein distributions or cell morphologies, while also assessing technical aspects such as artifacts and noise levels. The task requires MLLMs to perform anomaly detection and image comparison, going beyond simple perception in two key ways: analysis must consider sample preparation context, and expert knowledge is needed to evaluate biological features and technical artifacts. 

\paragraph{Hypothesis generation}
The next step is proposing mechanistic hypotheses to explain experimental data. For example, when a drug treatment causes mitochondrial fragmentation in cells, a hypothesis might be that the drug disrupted calcium homeostasis, activating the DRP1 protein through a multi-step pathway, leading to fragmentation. This requires \textit{abductive reasoning} \cite{douven2011abduction, bhagavatula2019abductive, hessel2022abduction, wang2024exploring} as one must select from many possible hypotheses given incomplete information. The process demands the MLLM to integrate relevant domain knowledge with experimental context and visual features to reason about which hypothesis best explains the observations.

\newcommand{\crossmark}{\ding{53}}
\newcommand{\halfmark}

\begin{table}[t]
\centering
\caption{Comparing scientific multimodal benchmarks close to MicroVQA for in terms of reasoning level or difficulty beyond college level. We show difficulty level, scientific domain, dataset source, and size. Compared to others, MicroVQA either has higher difficulty level, or it has higher reasoning level (which is established in \cref{fig:blooms}). Compared to others at the same reasoning level, namely MMMU-Pro, it has similar size.}
\label{tab:benchmark-comparison}
\resizebox{\columnwidth}{!}{%
\begin{tabular}{>{\raggedright\arraybackslash}p{2.7cm} 
                >{\centering\arraybackslash}p{2.0cm} 
                >{\centering\arraybackslash}p{2.0cm} 
                >{\raggedright\arraybackslash}p{2.9cm} 
                >{\centering\arraybackslash}p{1.0cm}}
\toprule
Benchmark & \shortstack{Difficulty \\ level}  & Domain & Source & Size \\
\midrule
\textbf{MicroVQA} & research & microscopy & expert-curated & 1,042 \\
MMSci~\cite{li2024mmsci} & research & science & paper figures & 7132 \\
LabBench~\cite{laurent2024lab} & research & biology & webQA & 181 \\
PathVQA~\cite{he2020pathvqa} & graduate & pathology & texbooks & 16.3k  \\
OmnimedVQA$^{*}$~\cite{hu2024omnimedvqa} & graduate & medical & classification dataset & 127.9k \\
Microbench~\cite{lozano2024mu} & graduate & microscopy & classification dataset & 17.2k \\
MMMU~\cite{yue2024mmmu} & undergraduate & general & textbooks, webQA & 11k\\
MMMU Pro~\cite{yue2024mmmu_pro} & undergraduate & general & MCQ dataset & 1,730\\
Science QA~\cite{lu2022learn} & grade-school & science & exams & 16.8k \\
\bottomrule
\end{tabular}%
}
\end{table}

\begin{figure}[t]
    \centering
    \includegraphics[width=\linewidth]{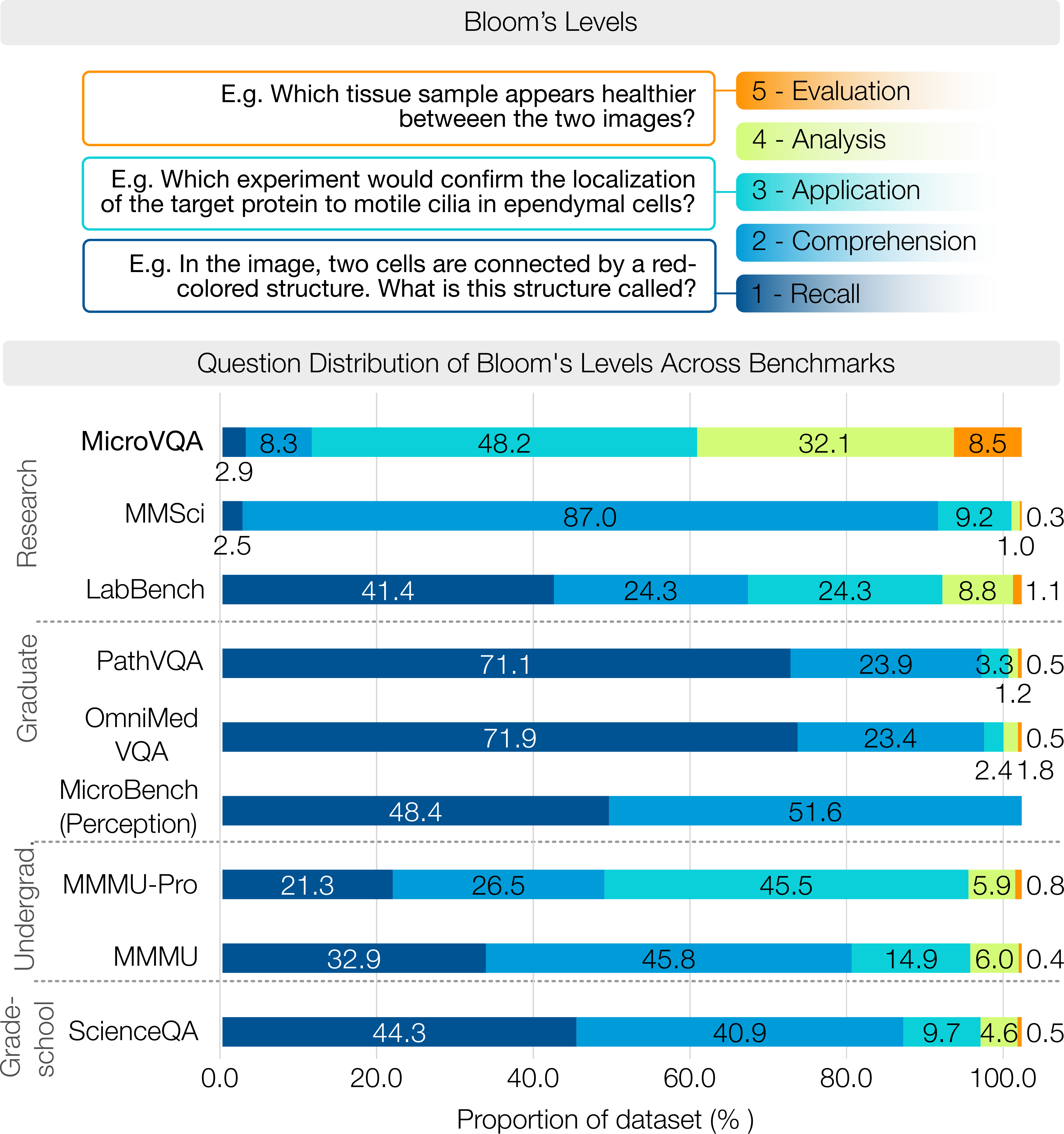}
    \caption{
    Composition of scientific MLLM benchmarks regarding estimated Bloom's taxonomy \cite{Bloom1956}. Higher levels are more cognitively challenging. MicroVQA has more questions at higher levels compared to other benchmarks, for example, MMMU \cite{yue2024mmmu} and ScienceQA \cite{lu2022learn},  while perception-driven medical benchmarks like OmniMedVQA are at lower levels.
    }
    \label{fig:blooms}
\end{figure}

\paragraph{Experiment proposal}
The final step is designing experiments to validate hypotheses. In microscopy, this often involves selecting appropriate assays and controls, requiring knowledge of suitable protocols and \textit{deductive reasoning} about whether proposed experiments will provide evidence for or against the hypothesis. This task also may involve adjusting a particular experiment to address technical issues, like how to ensure a good signal-to-noise ratio, and this requires both image understanding and knowledge of protocols.

In \cref{sec:human-reasoning}, we show an example question in each task with an expert-annotated `reasoning trace' -- it shows the type of reasoning required for typical questions.

\subsection{Analysis of MicroVQA benchmark}
\label{sec:benchmark-analysis}
Scientific discovery requires complex reasoning beyond basic perception and knowledge, and it requires expertise beyond school-level education. Here we show that MicroVQA addresses a gap that prior multimodal scientific benchmarks do not test \textit{high level reasoning} at \textit{research-level} difficulty.

To assess reasoning levels in MLLM benchmarks, we use Bloom's taxonomy \cite{Bloom1956}, which hierarchically classifies cognitive difficulty from recall to evaluation. We fine-tune GPT-4o to classify Bloom's taxonomy on related MLLM benchmarks with results in \cref{fig:blooms}; in \cref{tab:benchmark-comparison} we additionally show key attributes of the same benchmarks. \cref{fig:blooms} shows that benchmarks having research-level and graduate-level difficulty are dominated by lower reasoning questions (levels 1-2), while our MicroVQA has higher level reasoning (levels 3-4). This is not surprising since OmnimedVQA and MicroBench derive questions from prior classification datasets, while our task definition and data collection approach does reflect high-level reasoning.

On the other hand, some benchmarks have higher Bloom's reasoning level closer to MicroVQA -- MMMU, MMMU-Pro, and Science-QA --  but their difficulty level is lower -- undergraduate or grade-school. The most comparable benchmark for reasoning is the undergraduate-level MMMU-Pro, and its dataset size is 1,730 compared to our 1,042: this shows that for reasoning-intensive benchmarks, sizes beyond 1,000 are very high.

\section{MCQ generation process}
\label{sec:mcqgen}

\subsection{Creation of raw VQA samples by experts}
\label{sec:mcqgen-collection}

\begin{figure*}[th]
    \centering
    \vspace{-1em}
    \includegraphics[width=0.9\textwidth]{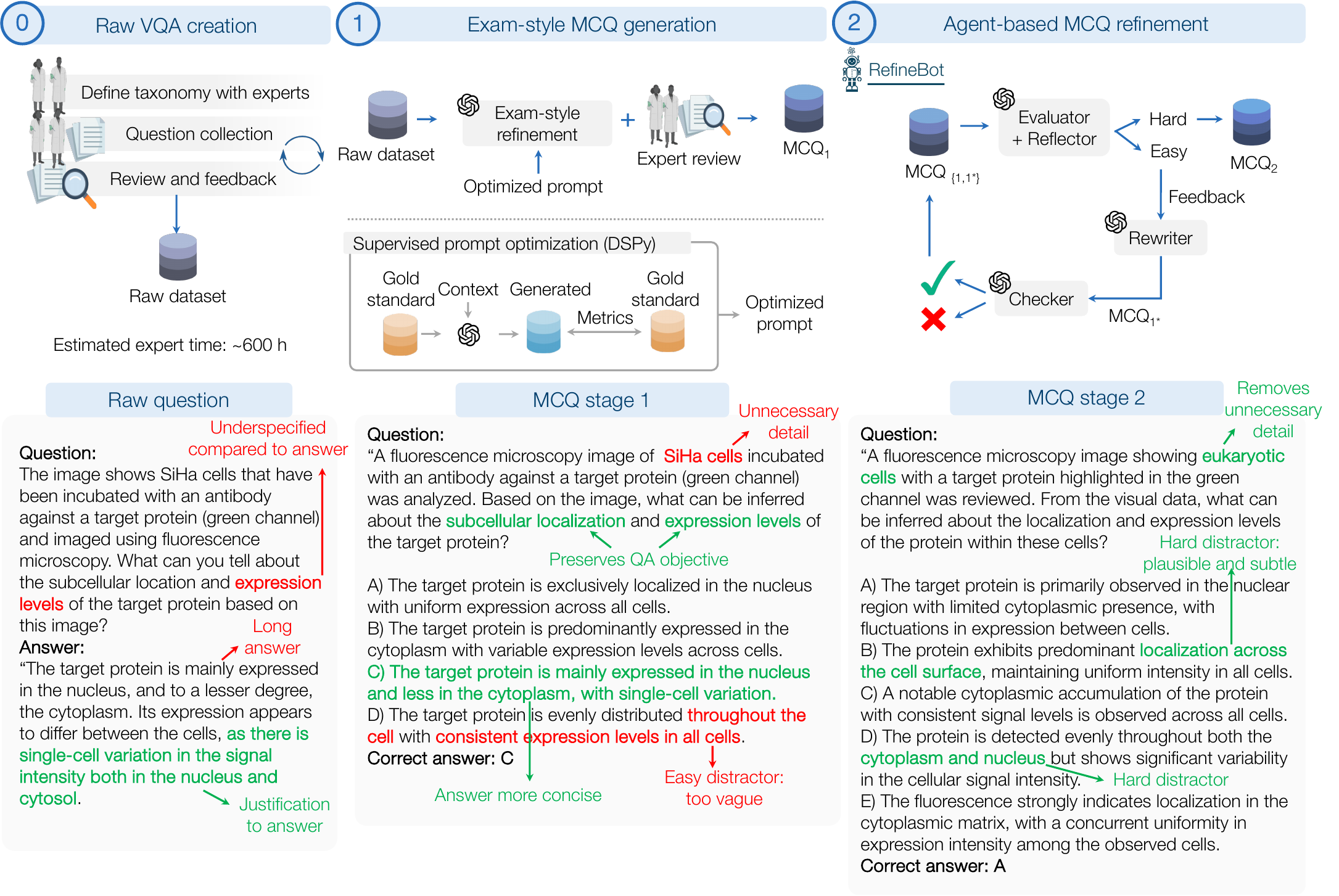}
    \vspace{-1em}
    \caption{Constructing the MicroVQA multiple choice questions. (0) We defined tasks with domain biological scientists and created 1,061 raw VQA samples. (1) The raw samples were aligned to an exam-style MCQ by manually transforming a small set and optimizing an LLM prompt to match that alignment. (2) MCQs are further improved using RefineBot, a new iterative method to make MCQs more challenging. The lower panel shows an example MCQ from raw VQA to final: the annotations highlight key changes that we further explore in \cref{sec:appendix-multichoice_gen}, where \red{red} indicates issues, and \color{Green} green \color{black} indicates good attributes.}
    \vspace{-1em}
    \label{fig:creation}
    
\end{figure*}

\paragraph{Expert generation of raw VQA samples}
Creating VQA triples suitable for research-level reasoning tasks is time-consuming and requires expertise. We recruited 12 human experts to each create approximately 90 VQA samples, taking about 30–40 minutes per sample. Each submission included an image set, a question, a detailed response, and contextual information like experimental context and motivation. They are raw VQA triples, $(v_0,q_0,a_0)$ --  `raw' means the question and answer can be much longer and more detailed than typical VQA benchmarks. This allowed expert annotators to include enough detail to match what is a valuable input and response for an MLLM. \cref{sec:appendix-question_collection} shows materials that prompt question creators. The images were sourced from the contributor's own experiments, image databases, or research papers published since 2024, and all image licenses permit redistribution. 

\paragraph{Quality control}
We had three quality control criteria: questions should be challenging and not solvable with undergraduate-level knowledge; they should be aligned with the defined tasks; and there should be diversity in image content. Questions were reviewed against these criteria for each contributor, and feedback was provided for further VQA samples. These review rounds occurred after submitting the first 10 questions, and the first 40 questions.

\subsection{MCQ generation from raw VQA samples}
\label{sec:mcqgen-method}
To benchmark MLLMs, we transform raw expert-created VQA samples to multiple-choice questions (MCQs) and propose new methods for MCQ generation.
\paragraph{Motivation: naive MCQ generation is inadequate}
 The standard approach for VQA benchmarks is to zero-shot prompt an LLM with raw VQA samples \cite{alhazmi2024distractor}, but this has two problems. The first is that generated MCQs do not follow established educational MCQ design principles (\cref{sec:appendix-multichoice_gen}), likely because the raw VQA questions and answers are often long and variable. Our stage 1 `exam alignment', which we describe in the next section, addresses this challenge. The second challenge is that generated MCQs do not properly test MLLM capabilities. Evaluating GPT-4o on these MCQs scores 93\% even \textit{without the image}, despite most of the dataset questions clearly requiring image analysis, and we investigate this by inspecting the chain-of-thought outputs in \cref{fig:creation} and \cref{sec:appendix-multichoice_gen}. Many distractors can be easily eliminated based on general biology knowledge, or because they are too vague compared to the correct option. We hypothesize that MCQ and distractor generation is challenging when the target answer string is long, and where the subject matter is specialized, suggesting that other scientific benchmarks may encounter similar issues. Overall, this motivates a more involved approach to distractor generation.

\vspace{-1em}

\paragraph{Stage 1: Exam-style MCQ generation}
Here we ensured that multiple-choice questions (MCQs) conformed to established design principles for biomedical exams (\cref{fig:creation}). This effort was led by a physician-scientist co-author trained in medical board exam item writing, who reviewed educational literature on MCQ design \cite{nbmeguidelines2021} and Bloom’s Taxonomy for assessing cognitive skills in biomedical exams \cite{revisedblooms2001, crowe2008-bloombiology, zaidi2017-bloomhistology}. We began by manually transforming 50 user-submitted raw question-answer pairs, $(q_0, a_0)$, into one-best-answer MCQs with $k$ distractors, $(q_1, a_1, {\mathbf{d_1}})$. These `Gold Standard' MCQs retained the original questions' meanings while strictly adhering to NBME standards \cite{nbmeguidelines2021} and minimizing cues that enable test-wise guessing. We used these gold-standard MCQs in a supervised learning setup to develop a general LLM prompt, $p$, that maps raw question-answer pairs to exam-style MCQs. We leverage the DSPy framework \cite{khattab2022demonstrate, khattab2023dspy}, which optimizes $p$ to match the gold-standard MCQ outputs. Additionally, the optimization process incorporated supervision from LLM-based quality metrics: content similarity, NBME-aligned formatting, and the absence of extraneous clues. Further details are in \cref{sec:appendix-multichoice_gen}.

\paragraph{Stage 2: RefineBot question refinement}
Although Stage 1 produces exam-aligned MCQs, many remain easily solvable due to language shortcuts (see \cref{sec:appendix-benchmark}). To enhance difficulty, we introduce \textit{RefineBot}, a method that increases MCQ complexity (\cref{fig:creation}). The key idea is that weaknesses in MCQ construction are revealed by the chain-of-thought (CoT) answers. The first LLM agent, the \textit{evaluator/reflector}, answers the MCQ with CoT and then reflects on the strategies used. For example, a question about processes inside a virus might have a distractor referring to surface processes -- this can be eliminated without referencing the image. The LLM summarizes its solution strategy into a reflection text: in this case, that the distractor was implausible based on knowledge that the stated process does not not occur inside the virus, and so does not match the question. It passes the reflection to the \textit{rewriter} LLM agent. The rewriter revises the question stem and generates new distractors to invalidate the identified strategies: in this example it may create distractors referring to processes that do occur inside a virus. To prevent significant changes to the question-answer pair over iterations, we employ an LLM \textit{checker} to ensure semantic consistency with the original pair. The revised MCQ is returned to the evaluator; if it still answers correctly, the loop continues. The process stops if the rewriter fails the \textit{checker} or after $n$ iterations. If it fails, RefineBot can be rerun with a different seed, often succeeding—a form of inference-time scaling \cite{brown2024large}. All agents are GPT-4o-0806, but to mitigate potential bias against 4o in final evaluation, we also use Claude-3.5-Sonnet-0620 as the \textit{evaluator} agent; we assess biases in the experiments section.

\noindent\textbf{Final MCQ quality check} Since  MCQ generation involves LLM processing, we need to verify the correctness of the final answer. Each
MCQ is manually reviewed by the same expert who created that particular question (between 80 and 120 per person). For any question with issues, that expert makes minimal changes to the question correct it.
\section{Experiments}
\label{sec:results}
\subsection{Benchmarking MLLMs with MicroVQA}
\label{fig:results-main}
We evaluate  state-of-the art multimodal large language models (MLLMs) on the MicroVQA benchmark. We include open and closed models spanning the categories: reasoning, large, small, and medical -- medical is the closest domain to microscopy with specialist MLLMs. We utilize standard chain-of-thought prompting \cite{yue2024mmmu_pro} (deatails in \cref{sec:appendix-results}). We report mean accuracy in multiple-choice VQA for each tasks \cref{tab:results}. We also perform and analyze no-image \cref{sec:appendix-language-ablations}.

\paragraph{MicroVQA is challenging for all MLLMs}
Our evaluation reveals a substantial gap between current MLLMs and the upper bound, with the leading model, o1, at 52.8. There is surprisingly little variation in performance between models, with most closed and open models all scoring above 40, with the lowest score (Llama-3.2-11b) likely due to instruction-following issues. A human baseline (\cref{sec:results-humanbaseline}) scores only 50.3 -- this shows that biology experts are specialized to subdomains, and are challenged by different subdomains. In fact this demonstrates that experts could benefit from MLLMs that can solve MicroVQA's tasks. The performance variation across the 3 tasks is small -- the highest score is 56.4 on  `expert visual understanding (V)', compared to the highest overall of 52.8. The `hypothesis generation' task is the hardest for all models and the gap is strongest for smaller models. For example the gap between hypothesis generation and expert visual understanding is 3.7 for the strongest large model, but 10.7 on the strongest small model.

\begin{table}[t]
    \centering
    \caption{VQA accuracy on MicroVQA by task: expert visual understanding (V), hypothesis generation (H),  experiment proposal (E). Models marked $^*$ were used in MCQ generation, which may affect comparative performance (see \cref{fig:results-main}). The model $^{\dagger}$ is the base LlaVA for LLaVA-Med. Best values are bolded. 
    }
    \label{tab:results}
    \resizebox{\linewidth}{!}{%
    \begin{tabular}{@{}p{1em}lcccc}
    \toprule
        & \textbf{Model} & \shortstack{\textbf{Overall}}& \shortstack{\textbf{V}} & \shortstack{\textbf{H}} & \textbf{E} \\
        \hline
        \multirow{1}{*}{\rotatebox[origin=c]{90}{\scriptsize \makecell{R}}} 
        & o1 \cite{jaech2024openai} & \textbf{52.8} & \uline{55.4} & \textbf{50.2} &\textbf{53.0} \\
        \hline 
        \multirow{9}{*}{\rotatebox[origin=c]{90}{\scriptsize \makecell{large models}}} 
         & $^*$Claude-3.5-Sonnet \cite{anthropic2024claude3} & \uline{51.7} & 54.1 & \textbf{50.2} & 50.4 \\
         & Gemini-Pro-1.5 \cite{team2023gemini} & 51.1 & 52.0 & \textbf{50.2} & 50.9 \\
         & Pixtral-Large \cite{agrawal2024pixtral} & 49.8 & 50.8 & 49.5 & 48.7 \\
         & Grok-2-Vision \cite{xai2024grok} & 48.4 & 50.3 & 46.4 & 48.7 \\
         & Qwen-2-vl-72b-Instruct \cite{wang2024qwen2} & 47.5 & 49.2 & 45.7 & 47.8 \\
         & VILA1.5-40b \cite{lin2024vila} & 47.5 & 47.2 & 47.9 & 47.4 \\
         & $^*$GPT-4o \cite{achiam2023gpt4} & 45.6 & 48.7 & 43.1 & 44.8 \\
         & Llama-3.1-Nemotron-70b-Instruct \cite{wang2024helpsteer2} & 44.2 & 44.9 & 43.3 & 44.8 \\
         & Llama-3.2-90b-Vision-Instruct \cite{dubey2024llama} & 42.4 & 44.9 & 42.1 & 38.7 \\
        \hline 
        \multirow{7}{*}{\rotatebox[origin=c]{90}{\scriptsize \makecell{small models}}} 
         & Qwen-2-VL-7b \cite{wang2024qwen2} & 48.8 & 54.1 & 43.3 & 49.6 \\
         & Claude-3.5-Haiku \cite{anthropic2024claude3} & 47.1 & 48.0 & 43.8 & \uline{51.7} \\
         & Gemini-Flash-1.5-8b \cite{team2023gemini} & 46.7 & 48.7 & 43.6 & 49.1 \\
         & GPT-4o-mini \cite{achiam2023gpt4} & 46.2 & 48.5 & 43.6 & 47.0 \\
         & Pixtral-12b \cite{agrawal2024pixtral} & 45.6 & 46.9 & 44.8 & 44.8 \\
         & VILA1.5-13b \cite{lin2024vila} & 41.8 & 41.8 & 47.5 & 40.9 \\
         & Llama-3.2-11b-vision-instruct \cite{dubey2024llama} & 30.3 & 32.4 & 29.3 & 28.7 \\
         \hline 
        \multirow{2}{*}{\rotatebox[origin=c]{90}{\scriptsize \makecell{medical}}} 
         & LLaVA-Med-Mistral-7B \cite{li2024llava} & 43.0 & 37.3 & 47.1 & 41.6 \\
         & $^{\dagger}$LLaVA-Mistral-7B \cite{liu2024visual} & 39.8 & 31.6 & 43.1 & 37.1 \\
         \hline
         \multirow{2}{*}{\rotatebox[origin=c]{90}{\scriptsize \makecell{}}} 
         & Random & 22.0 & 21.9 & 21.8 & 21.9 \\
         & Human & 50.3 & 52.7 & 47.5 & 51.4 \\
    \bottomrule
    \end{tabular}%
     }
\end{table}

\begin{table}[t]
\footnotesize
    \centering
    \caption{Ablation study on MicroVQA MCQ generation stages (shown in \cref{fig:creation}). Accuracy is high because MCQs have shortcuts (\cref{sec:mcqgen-collection}) after `Stage 1' exam alignment, but is lower after `Stage 2'. Final column is the relative decrease in accuracy. Models with $^*$ were used in Stage 2 generation and have the biggest accuracy drops (\textbf{bolded}). They are grouped with different models from the same provider in $^{**}$. 
    }
    \vspace{-0.5em}
    \label{tab:mcq_ablation}
\begin{tabular}{lccc} 
\toprule
& \multicolumn{1}{l}{\textbf{Stage 1}} & \multicolumn{1}{l}{\textbf{Stage 2}} & \multicolumn{1}{l}{\textbf{Relative drop}} \\
\hline
$^*$GPT-4o & 79.7 & 46.8 & -41.2 \\
$^{**}$GPT-4o-mini & 75.6 & 46.2 & -39.0 \\
$^*$Claude-3.5-Sonnet & 82.2 & 51.7 & -37.1 \\
$^{**}$Claude-3.5-Haiku & 77.3 & 47.1 & -39.0 \\
\hline
o1 & 81.6 & 52.8 & -35.3 \\
Pixtral-Large & 80.1 & 49.8 & -37.8 \\
Gemini-Pro-1.5 & 79.1 & 51.1 & -35.4\\
\bottomrule
\end{tabular}
\end{table}

\paragraph{Smaller models are surprisingly competitive}
For all models with a lower-parameter equivalent -- Gemini-1.5, QwenVL, VILA, and Llama -- the drop in performance due to size is less than 3 for all except Llama. Moreover, these drops are smaller than on other multimodal reasoning benchmarks like MMMU, where for example, Gemini drops by 9.6 between Pro and Flash-8B \cite{yue2024mmmu}. Typically, smaller models have the same vision encoder, but a smaller LLM \cite{wang2024qwen2}. Since the size of the language model has a small impact, we hypothesize that for the `solved' part of MicroVQA, the language-only reasoning is relatively simple compared to domains like math that require multi-step logic. This suggests that other aspects of MLLM reasoning are more challenging, like multimodal reasoning. If true, this suggests that future work could focus on stronger image representations, and this idea is supported by the qualitative error analysis in \cref{sec:results-errors}.

\paragraph{Specialist biomedical training does improve performance}
While no specialist MLLMs have been developed for microscopy research in particular, LLaVA-Med is  fine-tuned on scientific articles from PubMed that include data types overlapping with MicroVQA -- especially (tissue) pathology. In \cref{tab:results}, we compare LLaVA-Med against it's base model, LLaVA-Mistral-7B and find overall stronger performance by 4.5 points. This suggests two clear opportunities for improving MicroVQA performance. The first is to instruct-tune with a dataset that better aligns with the MicroVQA data domain: not only tissue pathology, but also fluorescence and electron microscopy, which is available in scientific articles \cite{zhang2023biomedclip}. The second is to simply begin training from a stronger open model like Pixtral. Beyond that, there is more to explore in MLLM specialization, such as tuning the base image encoder for microscopy \cite{lozano2025biomedica}. 

\begin{figure}[t]
    \centering
    \includegraphics[width=0.85\columnwidth]{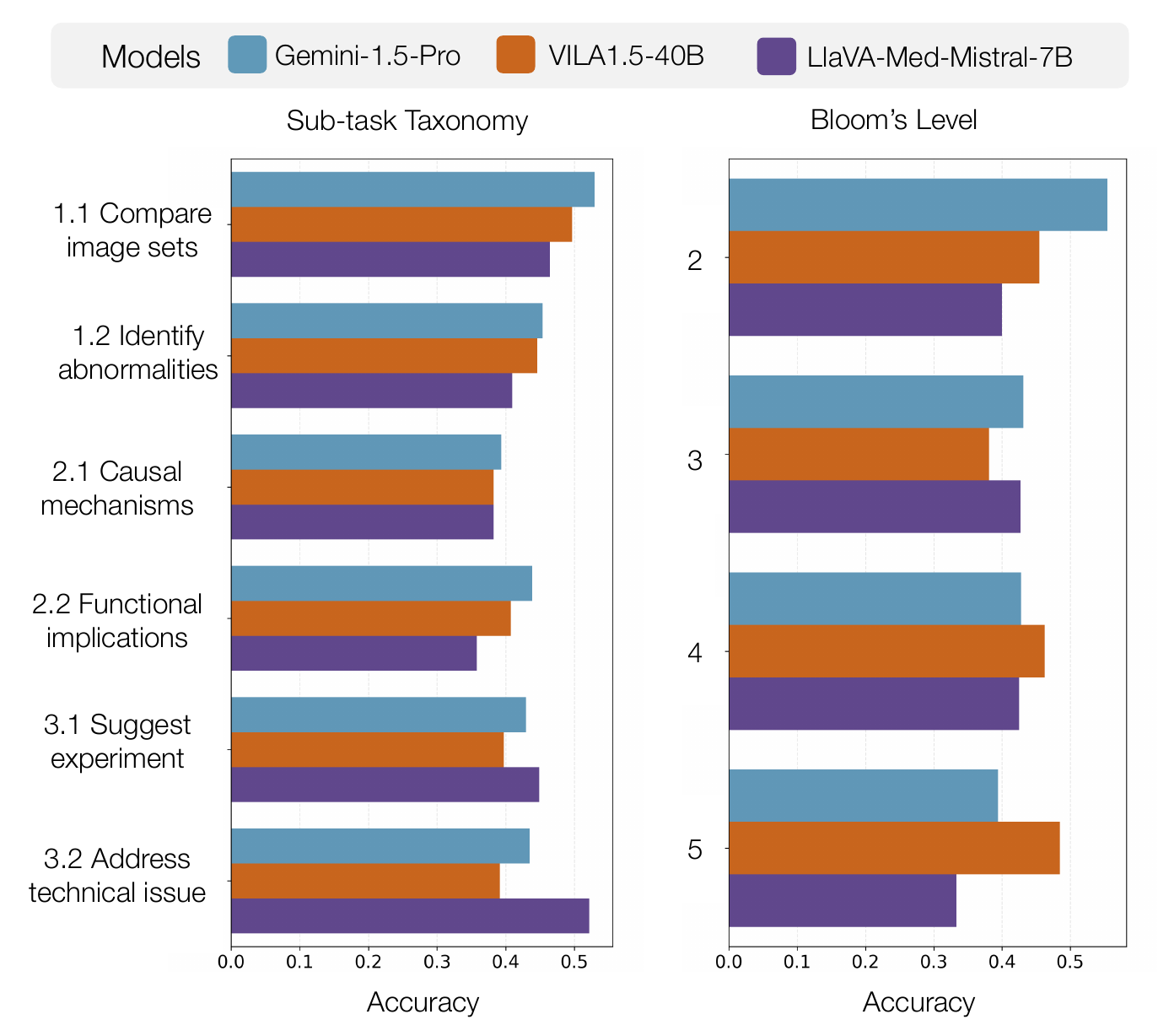}
    \caption{Performance by sub-task and Bloom's level for best models: Gemini-1.5-Pro (closed source), VILA1.5-40B (open-source), and LlaVA-Med-Mistral-7B (medical).}
    \vspace{-1em}
    \label{fig:analysis_modality}
\end{figure}

\paragraph{RefineBot MCQ generation is very effective but introduces small model biases} 
We introduced a new method for refining MCQs to make them more challenging -- RefineBot. \cref{tab:mcq_ablation} shows that between stage 1 and stage 2 (before and after RefineBot), the relative accuracy for all models drops by between 35\% and 42\%. Firstly, this validates that RefineBot is a valuable tool for making benchmarks harder. However, the most significant drops are for models used by RefineBot (GPT-4o \& Claude-3.5-Sonnet, and their smaller versions, GPT-4o-mini \& Claude-3-Haiku). The adversarial process introduced a small bias against these models (and we hypothesise the bias would be higher if using only one model instead of two). Despite making fair evaluation slightly more challenging, our use of frontier LLMs in the refinement process is well-motivated, as these models are best positioned to generate challenging, high-quality MCQs that probe the boundaries of current capabilities. Moreover, the bias seems modest -- less than 10 points of relative drop -- while the benefit is a huge increase in difficulty across all test MLLMs, as seen in \cref{tab:mcq_ablation}.

\paragraph{Metadata-based analysis supports deeper error analysis}
\cref{fig:analysis_modality} shows error rates based on attribute tag. In \cref{sec:appendix-results} we explore results further -- namely that multi-image reasoning has surprisingly good performance, higher level Bloom's questions are harder, and (not shown) rare image modalities perform worse.

\subsection{Error analysis on MicroVQA}
\label{sec:results-errors}
To understand model errors, three biomedical experts reviewed the Chain-of-Thought (CoT) responses of 30 samples on Claude 3.5 Sonnet, with each review requiring at least 45 mins of careful analysis. Errors were classified into a major category and minor contributing errors were also noted. Expert perception accounted for 50\% of errors, 30\% were knowledge errors, and 13\% were due to overgeneralization; the final 7\% were text hallucination or general reasoning error. \cref{sec:appendix-error-analysis} shows examples for the most common types. We expand on this analysis with an automated version across the complete dataset in \cref{sec:appendix-reasoning-analysis}.

\noindent\textbf{Expert perception errors} occurred when the model misinterpreted visual features, leading to incorrect inferences. For example, in \cref{fig:error_perception_00}, the model judges the small electron-dense objects as ribosomes, rather than stress granules --  ribosomes tend to be smaller, have a more regular shape, and lower contrast.  Language bias may have contributed to the preference for ribosomes since they are very commonly studied in EM literature available to LLM training. Overall, perception errors were the dominant error type (50\%), which suggests that future methods should improve vision representations in MLLMs for microscopy.

\noindent\textbf{Knowledge errors} highlight gaps in nuanced biomedical knowledge. One example  in \cref{fig:error_knowledge_01} deals with understanding how protein localization changes during cell signaling. An experiment gives information about a transmembrane signaling protein, and the question asks to interpret the image and explain the function in these cells. The model correctly perceives a punctate distribution, but assumes the role of coordinating intracellular vesicle trafficking. This represents a shallow interpretation that could have been improved with specialized biomedical knowledge. These errors could be mitigated by improving knowledge in MLLMs, either in training or as a database at inference.

\noindent\textbf{Overgeneralization errors} (and simplification errors) reflect the model's tendency to apply broad scientific principles without regard for specific context -- arguably they are \textit{reasoning} errors. 
In one example in \cref{fig:error_generalization_00}, the question is to compare the risk of malignancy of a tumor, schwannoma, to other nerve sheath tumors. During reasoning, the model proposes a simplified question -- what is true about schwannoma -- which it answers while ignoring the comparison.

\section{Conclusion}
\label{sec:conclusion}



MLLMs hold great potential for advancing scientific discovery. We aim to inspire the development of broadly applicable scientific AI systems. Future directions include training models for more robust image representations, integrating knowledge bases, exploring open evaluation methods, and studying LLMs’ reasoning over hypotheses. 
Beyond microscopy, we hope our benchmark serves as a blueprint for AI evaluation in expert-driven domains. More generally we aspire for different application domains -- biomedicine, chemistry, materials science -- to share methods towards the common goal of building scientific AI systems.

\FloatBarrier

{
    \small
    \bibliographystyle{ieeenat_fullname}
    \bibliography{main}
}
\appendix
\clearpage
\setcounter{page}{1}
\onecolumn
\maketitlesupplementary

Appendices A-C include acknowledgments, author contributions, ethics, and discussion of limitations and future work. The remaining sections correspond to sections in the paper. \cref{sec:appendix-benchmark} is benchmark details from the main \cref{sec:benchmark}. \cref{sec:appendix-question_collection} is MCQ generation details from the main \cref{sec:mcqgen}. \cref{sec:appendix-results} is the experiments details from the main \cref{sec:results}.

\vspace{-4em}
\part{}
{
\hypersetup{linkcolor=black}
\parttoc
}

\section{Acknowledgments and author contributions}
\textbf{Acknowledgements}
We thank Christian Kindermann for feedback on motivation for benchmarking and reasoning taxonomies; Siddharth Doshi for feedback on motivation for applications in scientific chat systems; and Krisjanis Mazurs for feedback on structuring user interviews. We also thank Assaf Zaritsky, Andrew S Moore, Pedro Guedes Dias, Angus Toland, Helen Healy, and Andrew Kassianos, and Loic Royer for discussions on defining the VQA tasks.

\textbf{Funding acknowledgements}
We gratefully acknowledge funding from HAI Hoffman-Yee (E.L., S.Y-L., J.M.H.B.), Schmidt Futures (E.L.), the Bridge2AI Program (NIH Common Fund; OT2 OD032742; E.L.,), Göran Gustafsson Foundation and Knut and Alice Wallenberg Foundation (KAW 2021.0346; E.L.) as well as to MU (HPA). S.C. received funding from NIH/NIGMS R35GM133460 and CZI A23-0264-001. J.N.H. was supported by a Postdoctoral Fellowship from EMBO Postdoctoral Fellowship (ALTF 556-2022).

\textbf{Author contributions}
\begin{itemize}
    \item Project conception: JB, JN, LBS, AL, SYL
    \item Defining the VQA tasks: JB, JN, JGG, WC, SC, JNH, CL, SYL
    \item Managing benchmark collection processes: JN, JB
    \item Creating benchmark questions: DB, ZC, SMH, AJ, WDL, MGN, RY, CZ, JNH, CL
    \item MCQ conversion pipeline: JB, JN, LBS
    \item Evaluations: JB, SRG, YS
    \item Qualitative error analysis: JN, JGG, CZ
    \item Quantitative error analysis: LBS
    \item Paper writing and figures: JB, JN, LBS
    \item Supervision: SYL, EL, CL, MDL, CNH, SC, WC
\end{itemize}

\iftoggle{cvprfinal}{
}{
This section is omitted for CVPR submission to adhere to blind review. 
}
\section{Ethics}
\textbf{Ethical use of biomedical data:} MicroVQA was developed with a commitment to ethical practices in handling biomedical research data. All microscopy images included in the dataset were either original, unpublished data or sourced from open-access articles (e.g., CC-BY licensed) published after January 2024. The dataset does not include patient-identifiable information, and MicroVQA adheres to applicable privacy and research ethics guidelines. The dataset does not constitute human subjects research.

\textbf{Potential societal impacts:} Biomedical AI models, including those trained on MicroVQA, have the potential to influence research and healthcare significantly, both positively and negatively. While these models can accelerate scientific discovery, they may also perpetuate biases in training data or result in uneven performance across demographic or biological groups. To mitigate these risks, MicroVQA includes metadata annotations to support the analysis of potential biases and emphasizes diverse data sourcing. We will engage with the research community to address emerging ethical concerns.

\textbf{Data licensing and usage:} MicroVQA is distributed under the Creative Commons Attribution-ShareAlike 4.0 International (\href{CC BY-SA 4.0}{https://creativecommons.org/licenses/by-sa/4.0/}) license to promote transparency and collaboration in the research community. This license reflects our efforts to balance open access with the ethical use of data, enabling both academic and commercial applications of the dataset while adhering to original license requirements. Full licensing details are available at \href{https://creativecommons.org/licenses/by-sa/4.0/}{https://creativecommons.org/licenses/by-sa/4.0/}.

\section{Limitations and future work}
While MicroVQA advances research-level reasoning benchmark significantly, we can identify a number of limitations and avenues for future work. 

\paragraph{Open vs closed evaluation} 
Like most VQA benchmarks, we used multiple-choice (MCQ) evaluation. However downstream applications will obviously operate in an open setting, without options and requiring more detail. MLLM research does have some open benchmarks \cite{ying2024mmt}, though impartial evaluation is a significant challenge. Our early experiments with open prompting suggest that models like GPT-4o tend to give very vague answers, and tend to depend strongly on the text part of the input prompt.

\paragraph{Dataset scale} Our final dataset had 1042 samples, which is smaller than some other VQA benchmarks like MicroBench \cite{lozano2024mu} and OmniMedVQA \cite{hu2024omnimedvqa}. MicroVQA's scale sufficient to compare overall scores between models, however statistical power starts to become limiting when trying to compare different models on subsets of the dataset. How does this compare to other benchmarks? As we can see from \cref{tab:benchmark-comparison} and \cref{fig:blooms}, the larger datasets tend to be those with lower-level reasoning according to Bloom's taxonomy. Their samples tend to be derived automatically from metadata tags or classification labels. Benchmarks with comparably high-level reasoning, like MMMU-Pro, have a similar scale to ours (1,700). In the language-only domain, the research-level reasoning benchmark GPQA has 448 samples. This trend reflects a clear, practical limitation: it is very time-consuming for humans to create difficult reasoning questions, and the pool of capable experts is small.

\paragraph{Breadth of dataset attributes in microscopy} We aimed for MicroVQA to cover a broad range of the microscopy discipline; for example, we have samples from all four relevant scales -- tissues, cells, subcellular, atomic.  However, each human annotator is an expert in a subset of microscopy and biology, so there are practical limits on dataset coverage. We, therefore, chose to focus on the most common imaging modalities --  brightfield, fluorescence, electron -- while skipping less common modalities like Raman spectroscopy. In terms of sample types, we emphasized those relevant to humans -- human and mouse -- while having only a few samples from rare organisms like bottlenose dolphin. 

\paragraph{Coverage of scientific reasoning} Looking even more broadly, the motivation of this work is to advance reasoning in scientific research generally. However, we use microscopy in particular. This is because of a clear trade-off in breadth vs depth. Our dataset strikes a good balance, given that microscopy represents a large portion of visual data in biomedicine and biology. Future work can reuse our framework to generate reasoning VQA benchmarks in other fields of visual biomedicine and biology, such as in medicine \cite{hu2024omnimedvqa} or ecology \cite{zhang2023biomedclip}.

\paragraph{Resolving MCQ shortcuts} A key contribution in this paper is the RefineBot for generating multiple choice questions (MCQs) that are too challenging to solve using text-only shortcuts. Our approach used the outputs of existing MLLMs as signal to make distractors harder. However, this cannot guarantee the removal of shortcuts; more generally, harder distractors may exist. Therefore, the next generation of LLMs may be able to identify and exploit new shortcuts. If that occurs, we can run RefineBot again using the new LLM, creating MicroVQA-2. This strategy could be used for other VQA evaluations. 

\section{Benchmark details}
\label{sec:appendix-benchmark}
\subsection{Accessing MicroVQA benchmark}
MicroVQA is an expert-curated benchmark for multimodal scientific reasoning, specifically in microscopy-based biological research. The dataset comprises 1,042 multiple-choice questions (MCQs) created by biological researchers to represent authentic scientific practice. Each MCQ is paired with an associated microscopy image and metadata describing the experimental context. A demonstration subset of 28 instances is available for initial exploration.

The MicroVQA dataset is publicly accessible on HuggingFace at \href{https://huggingface.co/datasets/jmhb/microvqa}{https://huggingface.co/datasets/jmhb/microvqa}. It is distributed under the Creative Commons Attribution-ShareAlike 4.0 International (CC BY-SA 4.0) license, allowing reuse and share-alike redistribution with attribution. The dataset is available in Apache Arrow and Parquet formats for efficient data handling. These formats support fast serialization, memory mapping, and streaming from the web, facilitating seamless integration into machine learning pipelines.

\subsection{Dataset Schema and Structure}
The dataset schema defines the fields included for each data instance. Figure \ref{fig:data_schema} provides a schema summary, while a sample data instance is detailed below \ref{fig:data_scema_instance}.

\begin{figure*}[htb]
\centering
\begin{mdframed}[backgroundcolor=json_bg, font=\small, leftmargin=3cm, rightmargin=3cm]
\scriptsize
\begin{verbatim}
{
    "image_id": datasets.Value("string"),
    "image": datasets.Image(decode=True, id=None),
    "label": datasets.ClassLabel(),
    "label_name": datasets.Value("string"),
    "domain": datasets.Value("string"),
    "subdomain": datasets.Value("string"),
    "modality": datasets.Value("string"),
    "submodality": datasets.Value("string"),
    "stain": datasets.Value("string"),
    "microns_per_pixel": datasets.Value("string"),
    "questions": datasets.Value("string"),
}
\end{verbatim}
\end{mdframed}
\caption{Example data schema.}
\label{fig:data_schema}
\end{figure*}

\begin{figure}[htb]
\centering
\begin{mdframed}[backgroundcolor=json_bg, font=\small, leftmargin=2cm, rightmargin=2cm]
\scriptsize
\begin{verbatim}
{
    "image_id": "d54bb7ec-284f-4218-a47d-af87bb371de5",
    "image": datasets.Image(decode=True, id=None),
    "label": datasets.ClassLabel(),
    "label_name": "pathology",
    "domain": "pathology",
    "subdomain": "gastrointestinal pathology",
    "modality": "light microscopy",
    "submodality": "brightfield microscopy",
    "stain": "H&E",
    "microns_per_pixel": "Not provided",
    "questions": "Question: 
        ```A hematoxylin and eosin (H&E)-stained micrograph shows cellular
        formations within a gastrointestinal tissue sample with unique 
        organization and cellular patterns. What pattern and morphology do these 
        cells most closely align with?
        
           A) Bundles of elongated cells with pale pink cytoplasm and uniform oval 
              nuclei in a crisscross pattern
           B) Circular groups of cells with mucin in central spaces and small,
              dark nuclei
           C) Dense arrays of polygonal cells with small nucleoli and vacuolated 
              cytoplasm
           D) Rows of cells with conspicuous cytoplasm and rounded nuclei
           E) Diffused arrangement of cells with transparent cytoplasm 
              and irregularly clustered chromatin
        
        Correct Answer: A) Bundles of elongated cells with pale pink cytoplasm  
        and uniform oval nuclei in a crisscross pattern```
}
\end{verbatim}
\end{mdframed}
\caption{Example data instance.}
\label{fig:data_scema_instance}
\end{figure}

\subsection{Training contamination mitigation}
To mitigate contamination, all data in MicroVQA were sourced from original, unpublished microscopy images or open-access articles (e.g., CC-BY-licensed) published after January 2024. Additionally, canary strings were embedded in the dataset to detect potential contamination during fine-tuning or model evaluation.

\subsection{Benchmark attribute details}
The MicroVQA benchmark is designed to evaluate multimodal reasoning capabilities in biomedical research contexts. The dataset incorporates diverse attributes such as organism, research subject, imaging modalities, and biological length scales (image scales) to ensure a comprehensive assessment\ref{tab:benchmark_attributes}. These metadata not only enhance the scientific relevance but also enable analysis of reasoning across different biological and microscopy contexts. This section describes the methodology used to define, compute, and annotate these key attributes.

\begin{figure}
    \centering
    \includegraphics[width=\linewidth]{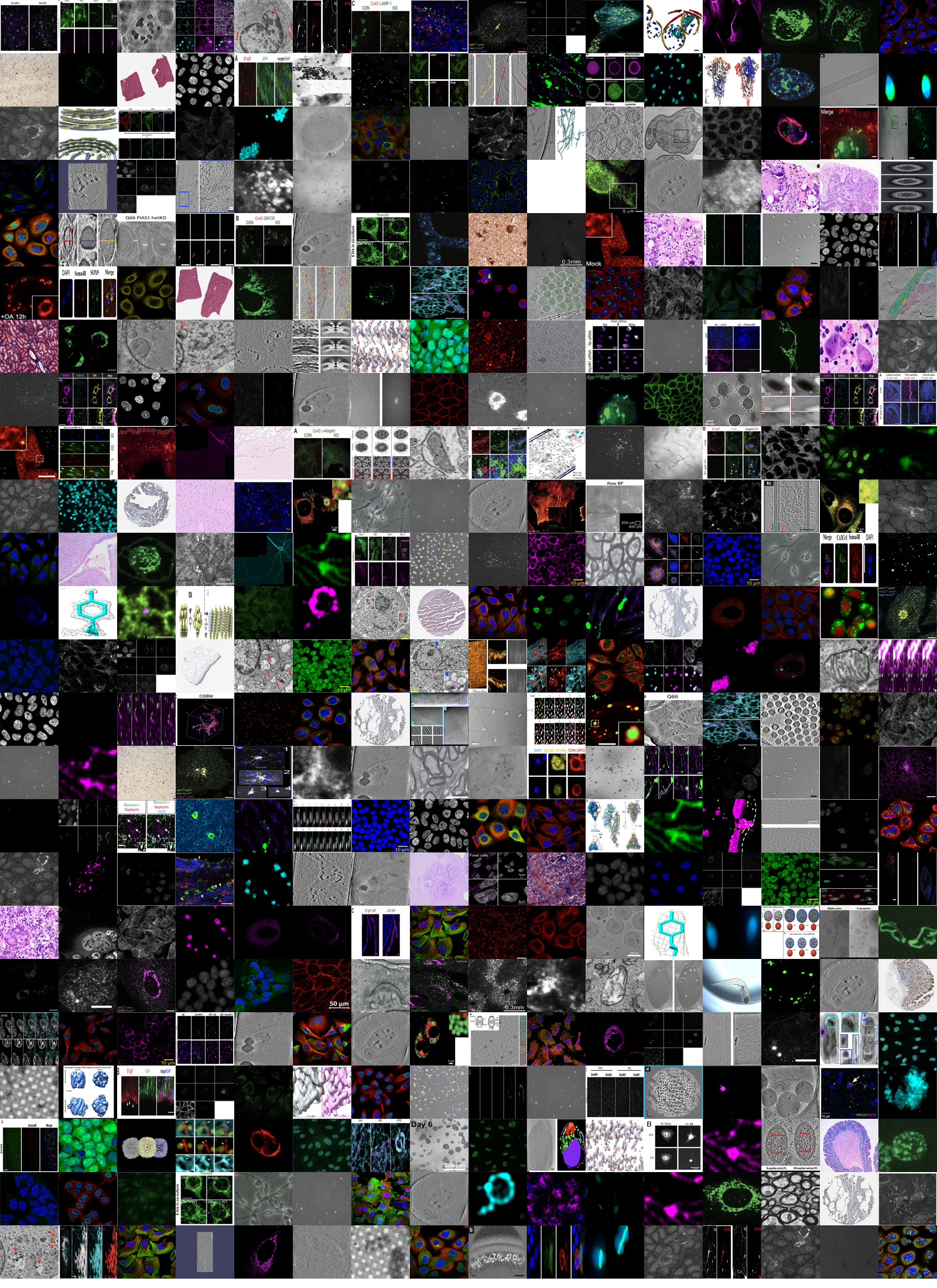}
    \caption{Collage of images from MicroVQA.}
    \label{fig:collage}
\end{figure}

The benchmark comprises 1,042 total questions, including 423 questions containing multiple images combined into panels. A collage of sample images is shown in \ref{fig:collage}. The average MCQ token lengths were computed using the \verb|o200k_base| tokenizer, with separate calculations for the raw input and formatted multiple-choice questions. Imaging modalities are broadly categorized into light, fluorescence, and electron microscopy. Biological length scales span four levels: tissue (millimeter scale), cellular (micrometer scale), subcellular (nanometer scale), and atomic (angstrom scale), enabling reasoning across diverse structural hierarchies.

To annotate attributes such as organism, specimen, and research subject, we manually reviewed random samples from the dataset and created a taxonomy to cover the most likely categories. A biomedical domain expert manually labeled these categories for 50 raw input question-answer pairs and performed initial prompt engineering. We used DSPy to optimize the prompt and few-shot examples on the labeled dataset. The input context included the taxonomy of organisms\ref{fig:organism_taxonomy} and research subjects \ref{fig:research_taxonomy}.
\begin{figure*}[htb]
\centering
\begin{mdframed}[backgroundcolor=prompt_bg, font=\small, leftmargin=2cm, rightmargin=2cm]
\scriptsize
\begin{verbatim}
    Taxonomy of organisms in biology| # Taxonomy of Organisms in Biology 
    and Biomedical Research:
    Apis mellifera (Honeybee)
        Social insect with complex behavior.
        Keywords: honeybee, social behavior, ecology, pollination.

    Arabidopsis thaliana (Thale Cress)
        Model plant organism.
        Keywords: Arabidopsis, plant genetics, photosynthesis.

    Caenorhabditis elegans (C. elegans)
        Transparent nematode worm.
        Keywords: worm, development, aging, programmed cell death.

    Danio rerio (Zebrafish)
        Fish with transparent embryos.
        Keywords: zebrafish, vertebrate development, embryology.
    ...

    Drosophila melanogaster (Fruit Fly)
        Insect with rapid life cycle.
        Keywords: fruit fly, genetics, development.
    
    Escherichia coli (E. coli)
        Bacterium used in genetics and molecular biology.
        Keywords: bacteria, gene expression, plasmids.

    ...
    Homo sapiens (Human Cells)
        Cultured human cells (e.g., HeLa, HEK293, A431, U2OS).
        Keywords: human cell lines, immortalized cell lines

    In vitro (None)
        Purified molecules and reactions in controlled environments. 
        No living cells or organisms.
        Keywords: purified proteins, purified molecules, cell-free systems
    ...

\end{verbatim}
\end{mdframed}
\caption{Examples of taxonomy classes used as context to LLM to assign an organism to a question. A YAML file with the full taxonomy will be released with the code.}
\label{fig:organism_taxonomy}
\end{figure*}

\begin{figure*}[htb]
\centering
\begin{mdframed}[backgroundcolor=prompt_bg, font=\small, leftmargin=2.5cm, rightmargin=2.5cm]
\scriptsize
\begin{verbatim}
    Taxonomy of research subjects| #Taxonomy of Research Subjects in Biology:
    Anatomy
        Study of the structure of organisms and their parts.
        Keywords: organs, tissues, morphology, physical organization.
    
    Biochemistry
        Chemical processes within living organisms.
        Keywords: proteins, enzymes, nucleic acids, metabolic pathways.
    ...    
    
    Cell and Molecular Biology
        Structure and function of cells and their molecules.
        Keywords: cellular processes, gene expression, molecular pathways.
    
    ...    
    Developmental Biology
        Process by which organisms grow and develop.
        Keywords: embryology, morphogenesis, genetic regulation.
    ...
    Genetics
        Study of heredity and gene function.
        Keywords: DNA, inheritance, genetic variation, gene expression.
    
    Immunology
        Immune system in health and disease.
        Keywords: antibodies, immune response, pathogens, vaccines.
        
    Microbiology
        Study of microorganisms.
        Keywords: bacteria, viruses, fungi, parasites, microbial ecology.
        
    Neurobiology
        Structure and function of the nervous system.
        Keywords: neurons, brain, neural circuits, cognition.
    ...    
    Structural Biology
        Molecular structure of biological macromolecules.
        Keywords: protein folding, nucleic acid structures, crystallography.
    ...
\end{verbatim}
\end{mdframed}
\caption{Examples of taxonomy classes used as context to LLM to assign a research subject to a question. A YAML file with the full taxonomy will be released with the code.}
\label{fig:research_taxonomy}
\end{figure*}

These annotations were the gold standard to guide DSPy prompt engineering for automated attribute labeling on the full dataset. Prompt instructions included fallback rules (e.g., tagging "None" for indeterminate attributes) to ensure accuracy and minimize hallucination. The optimized pipeline annotated all questions while ensuring the validity of the dataset attributes.

\subsection{Related benchmarks}
In section \ref{sec:benchmark-analysis} and \cref{tab:benchmark-comparison} we compare MicroVQA to related scientific multimodal benchmarks, focusing on its intersection of high-difficulty and advanced reasoning questions. Beyond these key attributes, we examine the sources of the questions. Most benchmarks rely on existing question repositories, such as exams, web-based QA datasets, textbooks, or other MCQ datasets. This approach bypasses the need for MCQ generation methods, as the questions and options are pre-tested and curated, making these benchmarks more representative of established tasks. Another common strategy involves reformatting classification datasets into MCQs using question templates, which increases task diversity but often limits scenario uniqueness and language variability. In contrast, MicroVQA is built by directly querying experts for challenging, domain-specific research questions. By starting with long form questions and answers, our approach requires an innovative MCQ generation method, ensuring uniquely tailored questions that push the boundaries of domain-specific reasoning and knowledge assessment.

\subsection{Bloom's taxonomy}
Bloom's Taxonomy is a model for evaluating cognitive skill levels in educational assessments. It has six hierarchical levels of learning: recall, understand/comprehend, apply, analyze, evaluate, and create. Multiple-choice questions can test five of Bloom's levels but not the sixth level, ``create''. These levels provide a structured framework for assessing lower-order and higher-order thinking skills, which are essential for designing robust benchmarks like MicroVQA.

Initial attempts to use LLMs to classify Bloom's levels for biomedical visual question answering (VQA) benchmarks revealed a systematic overestimation of cognitive skill levels. This discrepancy limits the naive use LLMs for automated Bloom's classification, perhaps due to the nuances of applying Bloom's taxonomy in domain-specific contexts. To ensure accurate classification of Bloom's level, a physician-scientist with NBME training in item writing and familiarity with Bloom's classification manually labeled Bloom's level for 200 examples from public MCQ questions, MicroVQA, MicroBench, and ScienceQA. These examples were used to fine-tune GPT-4o-mini (gpt-4o-mini-2024-07-18) to classify Bloom's level and align with expert judgments.

The fine-tuned Bloom's classifier was applied to MicroVQA and numerous public science/biomedical benchmarks to assess the cognitive skills evaluated by each question. Each question was mapped to one of Bloom's levels, providing insights into the distribution of cognitive tasks across the benchmark. This analysis revealed that while existing benchmarks predominantly tested lower-order skills (e.g., recall and comprehension), MicroVQA has a greater emphasis on questions that require higher-order reasoning (e.g., analyzing, evaluating), filling a critical gap in multimodal scientific reasoning assessments.

\subsection{Methodology for proposing tasks}
Here, we expand on the task proposal methodology introduced in \cref{sec:benchmark-tasks}. To define specific reasoning tasks aimed at advancing biomedical research, we conducted a comprehensive expert consultation process. Through structured interviews with nine Principal Investigators and postdocs specializing in microscopy research, including participants from both academia and industry, we developed a framework for tasks that would meaningfully integrate multimodal large language models (MLLMs) into microscopy research workflows.

The task selection process followed specific criteria: tasks must utilize image-text inputs and text outputs suitable for MLLMs; they should require higher-order reasoning beyond simple image processing; and they should emphasize core experimental activities (experimentation, analysis, and action) rather than auxiliary tasks like literature review or writing \cite{hope2023computational}.

Our methodology involved a two-phase interview process. The first phase consisted of initial one-hour individual sessions with eight research groups featuring open-ended, unstructured discussions about key challenges in microscopy experiments, potential applications of MLLMs, current workflows and unmet needs, and possible AI system integration points. One week later, we then conducted follow-up interviews to consolidate findings and build consensus.

To avoid anchoring biases, we intentionally withheld predefined categories during initial interviews. This approach enabled unbiased task identification and allowed us to capture a broad spectrum of potential use cases. Through this iterative process, we identified three essential capabilities that align with the scientific life cycle for microscopy experiments: advanced image interpretation, hypothesis generation, and experimental design/hypothesis testing.

\subsection{Guidance for question creators}
Question creators were given detailed guidance as pdfs, that we share at  \href{https://github.com/jmhb0/microvqa}{https://github.com/jmhb0/microvqa} under folder \texttt{benchmark/human\_collection}. During the submission process, we completed two rounds of quality control to ensure difficulty and alignment to the prompts, as described in \cref{sec:mcqgen-collection}.

\subsection{Microscopy image representations}
MLLMs are trained primarily on three-channel RGB images, however microscopy images are more diverse. To ensure alignment with how common MLLMs process images, we required that each image artifact for a single question was represented as no more than six RGB images. Where multiple images were needed, they could be multiple image files, or a single image that is concatenated. For videos, users could submit up to six frames; for 3d, they could submit slices or z-projections. 

An important difference is multichannel images. Many fluorescence microscopy images are multichannel, where the image collection ensures that each channel represents a certain structure. For example channel 0 may be nucleus, and channel 1 may be mitochondria. There are a few approaches. Some questions represent this situation as separate grayscale images for each channel. Others show a `composite image', which is a single image where one each channel is mapped to one of the RGB channels; e.g. a common visualization will put nucleus in blue and mitochondria in green. Some questions do both: grayscale images for each channel, along with a composite image merging them. 

Finally, we allowed screenshots from research papers or image software, since this is a realistic way that future LLM-users would use AI tools.

\section{MCQ generation details}
\label{sec:appendix-question_collection} 
\subsection{Question collection process details}
Data collection used a structured submission form shared with collaborators. The form required users to provide the following:
\begin{itemize}
    \item  Image or Image Set: Users uploaded microscopy images, optionally including multiple images for comparative questions.
    \item Experiment Details: A brief description of the experimental setup, imaging modalities, and any relevant biological context.
    \item Question Text: Each question was aligned with one of three predefined use cases: identifying interesting features, explaining mechanisms, or proposing next steps.
    \item Answer and Distractors: Contributors provided a correct answer and biologically plausible incorrect answers.
\end{itemize}

Quality control was performed in multiple rounds. To familiarize users with the task and data requirements, each user submitted an initial form with 4-6 questions for manual quality control and feedback. Feedback was provided to users on improving question specificity, ensuring the questions required image interpretation rather than relying solely on general knowledge. Contributors were encouraged to refine their submissions based on the feedback to better align with the tasks and goals of the MicroVQA benchmark.

Users were encouraged to submit data from multiple sources, imaging modalities (e.g., fluorescence, electron microscopy), and biological scales (e.g., tissue, cellular, subcellular) to promote diversity. Contributors could upload microscopy images from their experiments or use CC-BY publicly available images. Common image sources included preprints or publications released after January 2024 (to mitigate contamination), or from established image repositories \cite{thul2017subcellular, cho2022opencell, viana2023integrated, clark2013cancer, hein2025global, burgess2024orientation}. This ensured that all images were either original or sourced from open-license publications, enabling their use in an open-source benchmark.

\subsection{Experiment on naive MCQ generation}
\label{sec:appendix-multichoice_gen}
In \cref{sec:mcqgen-method}, we discuss how the naive approach to MCQ generation -- simply zero-shot prompting GPT -- leads to MCQs that are very easy for frontier models to solve. Which ultimately motivated our two-stage MCQ generation approach. Here, we discuss those experiments in more detail. We create a naive version of MicroVQA using a standard prompt \cite{hu2024omnimedvqa,wang2024mmlu}: 
\begin{mdframed}[backgroundcolor=prompt_bg, font=\small, leftmargin=0cm, rightmargin=0cm]
\scriptsize
\begin{lstlisting}
You are an expert in molecular and cell biology, and in microscopy.

I will give you an original biology-related question and its answer, your task is to rephrase an equivalent 
question with identical answer. The question related to an image, and we don't show the image.
Meanwhile, I want to transfer this QA-pair into a multi-choice question. Please generate 5 incorrect options to
construct the candidate options.

{{QUESTION}}

{{ANSWER}}

\end{lstlisting}
\end{mdframed}

Then, we evaluated this naive MicroVQA on the top closed source models with the image, and to study the over-reliance on language shortcuts, we evaluated a text-only version. In the text-only setting, we add the following to the prompt: ``If an image is mentioned, ignore this information and try your best to answer the question.''. \cref{tab:naive-performance} shows the results of this experiment.

\begin{table}[ht]
    \footnotesize
    \centering
    \caption{Performance on MicroVQA MCQs with naive MCQ generation. We report overall VQA and VQA without the image.}    
    \label{tab:naive-performance}
\begin{tabular}{lcc}
\toprule
  & \multicolumn{1}{l}{\textbf{VQA}} & \multicolumn{1}{l}{\textbf{VQA-no-image}} \\
  \hline
GPT-4o & 85.1 & 82.7 \\
Claude-3.5-Sonnet & 91.4 & 88.4 \\
Gemini-1.5-Pro & 88.5& 82.4 \\
\bottomrule
\end{tabular}
\end{table}

All models score over 80\% on questions that are from a specialized domain. We hypothesize that this reflects a problem with the non-trivial task of MCQ generation from raw VQAs for several reasons. First, the high `VQA-no-image' scores across all models suggest that selecting the correct option does not require image understanding, despite the raw VQA samples being designed to incorporate image-based reasoning. An alternative explanation is that the questions themselves may be too easy for advanced models. Second, we test this alternate explanation by performing open evaluation on a subset of the raw VQA queries-- that is, we queried GPT-4o with the questions but without the options. Here, the models rarely gave good responses, even after careful prompting. This suggests that models are leveraging test-taking strategies to eliminate distractors \cite{gierl2017developing}. Third, we examined the chain-of-thought responses in the VQA-no-image case, which suggested the models often eliminate distractors based on information from the text question. Therefore, we conclude that there are issues with the improper generation of the MCQs using the naive strategy.

What exactly are the problems with MCQ generation? Qualitatively, the types of errors are the same as those we discuss in \cref{sec:appendix-stage2-analysis}, so we refer readers to that section to get a deeper understanding.

\subsection{Stage 1 MCQ generation (exam alignment)}
The first stage transformed raw, open-ended questions and long-form answers into well-structured MCQs aligned with established principles of biomedical exam design. A physician-scientist trained in medical board exam item writing initiated the process by reviewing biomedical educational literature, NBME guidelines \cite{nbmeguidelines2021}, and Bloom’s taxonomy for assessing cognitive skills \cite{revisedblooms2001, crowe2008-bloombiology, zaidi2017-bloomhistology}. The expert manually converted 50 raw question-answer pairs into one-best-answer MCQs with biologically relevant distractors using the literature review and their prior training. This manual effort ensured that the MCQs preserved the original question's intent and scientific validity while adhering to educational standards. Distractors were crafted to require reasoning and prevent reliance on superficial cues like word association or general knowledge, thus maintaining the depth and complexity of the original content.

To scale this process, the expert designed an initial LLM prompt for automated Stage 1 MCQ generation. This prompt and the manually curated training data were input for DSPy-based prompt optimization. Using the \texttt{MIPROv2} optimizer, the prompt was refined within a Chain of Thought (CoT) framework combined with retrieval-augmented generation (RAG) pipelines powered by \texttt{o1-mini}. Optimization was guided by a custom metric evaluated by a separate LLM judge (\texttt{o1-mini}). The judge assessed each revised MCQ across three criteria: (1) semantic similarity to the original question and answer, (2) adherence to NBME formatting standards, and (3) minimization of extraneous information that could provide unintended clues. The optimization metric was weighted to prioritize preserving the original meaning and scientific validity, with lesser emphasis on formatting and extraneous text. Additional implementation details are provided in the source code.

\subsection{Stage 2 MCQ generation (RefineBot) details}
The goal of RefineBot is to take MCQs that are too easy for MLLMs -- especially where they are leveraging language-only shortcuts -- and to rewrite them to make them more challenging. It is the third panel in \cref{fig:creation}. For more details on what types of improvements are achieved, see \cref{sec:appendix-stage2-analysis}. The complete code for RefineBot is released.

\paragraph{A single RefineBot iteration}
Below, we show the pseudo-code for a single-question revision. The input is a question string, a list of options, and the index of the correct answer in the options. It returns a tuple \verb|(code, MCQ)| where \verb|code| is a string on whether the run was successful, and \verb|MCQ| is either the revised multiple choice question if the rewrite was successful or otherwise \verb|None|.

\begin{lstlisting}[
    basicstyle=\ttfamily\footnotesize,
    breaklines=true,
    columns=flexible,
    frame=single,
    backgroundcolor=\color{gray!10},
    numbers=left,
    numberstyle=\tiny,
    numbersep=5pt,
    showstringspaces=false,
    mathescape=true,
    xleftmargin=2em,
    framexleftmargin=1.5em
]
def revise_mcq(question: str, choices: list, correct_idx: int, max_iters: int = 5):
    # Store original version
    original = {'question': question, 'choices': choices, 'correct_idx': correct_idx}
    reflections = []
    
    for iter in range(max_iters):
        # Try answering without image
        result = evaluate_without_image(question, choices, correct_idx)
        
        if not result.can_answer:
            code = 'SUCCESS_NO_CHANGE' if iter == 0 else 'SUCCESS_REWRITE'
            return code, (question, choices, correct_idx)
            
        if iter == max_iters - 1:
            return 'FAIL_ITERATIONS', None
            
        # Reflect on strategies used to solve
        reflection = analyze_why_answerable(result)
        reflections += reflection

        # Rewrite based on all reflections from all iterations
        new_version = rewrite_question(reflections, original)
        
        # Verify the rewrite didn't change the meaning meaning preserved
        if not check_meaning_preserved(original, new_version):
            return 'FAIL_REWRITE', None
            
        # Update current version
        question = new_version['question']
        choices = new_version['choices']
        correct_idx = new_version['correct_idx']
    
    return 'FAIL_UNKNOWN', None
\end{lstlisting}

This process is also described in the main text. One detail to note: the \verb|rewrite_question| function receives the reflections from all prior iterations, not just the current one. In this way, the rewriting function can consider all possible strategies that have been identified. All LLM calls are to \verb|gpt-4o-2024-08-06|.

To better illustrate the method, we step through a single example. This way, we can show the prompts and understand responses.
\begin{enumerate}
    \item The evaluation of a question with the \verb|evaluate_without_image| prompt and response:
    
\begin{lstlisting}[
    basicstyle=\ttfamily\scriptsize,
    breaklines=true,
    columns=flexible,
    frame=single,
    backgroundcolor=\color{prompt_bg},
    showstringspaces=false,
    mathescape=true,
    xleftmargin=2em,
    framexleftmargin=1.5em
]
Prompt
--------------------------------------------------------------------------------
The following question is supposed to be paired with an image. We will not provide the image, so answer to the best of your ability.
In an electron microscopy image of a mouse primary embryonic cortical neuron expressing full-length human mHTT (BACHD), "eye-ball" shaped features are observed in the top left corner that appear anomalous. What is the most likely cause of these features?
Think step by step and then output the answer in the format of "The answer is (X)" at the end.

(a) Aggregation of human mHTT protein within the neuron
(b) Ice contamination from liquid nitrogen during sample preparation
(c) Formation of intracellular vesicles during neuron development
(d) Artifacts arising from excessive heavy metal staining

--------------------------------------------------------------------------------
Response (target answer is b)
--------------------------------------------------------------------------------
To determine the cause of the "eye-ball" shaped features observed in the electron microscopy image, we can analyze each option step by step:

(a) Aggregation of human mHTT protein within the neuron: Aggregations of proteins can indeed appear as anomalous structures under electron microscopy. However, these typically have irregular shapes and may not consistently resemble "eye-ball" shapes.

(b) Ice contamination from liquid nitrogen during sample preparation: Ice contamination can occur during sample preparation, particularly if samples are not handled properly when flash-frozen. This can lead to anomalous features appearing in microscopy images. These features might appear as circular or irregular shapes, potentially looking like "eye-balls."

(c) Formation of intracellular vesicles during neuron development: Intracellular vesicles are normal structures within neurons and other cells. Although they can appear during development, they are typically well characterized and documented, and are less likely to be described as anomalous in the context given.

(d) Artifacts arising from excessive heavy metal staining: Over staining can indeed produce artifacts in electron microscopy images. While staining artifacts are plausible, they might not specifically result in "eye-ball" shapes unless the staining process consistently targets specific round structures.

Considering the nature of the descriptors ("eye-ball" shaped, anomalous), option (b) ice contamination is often a source of unexpected features in microscopy images that could fit the appearance described, as ice can form circular shapes.

Therefore, the most likely cause of these features is (b) Ice contamination from liquid nitrogen during sample preparation.

The answer is (b).
\end{lstlisting}

\item If the answer is incorrect, then the question is already hard enough, and we can exit. Otherwise, we call the \verb|analyze_why_answerable| function, which reflects on how this answer was achieved. It continues in the same conversation:

\begin{lstlisting}[
    basicstyle=\ttfamily\scriptsize,
    breaklines=true,
    columns=flexible,
    frame=single,
    backgroundcolor=\color{prompt_bg},
    showstringspaces=false,
    mathescape=true,
    xleftmargin=2em,
    framexleftmargin=1.5em
]
--------------------------------------------------------------------------------
User
--------------------------------------------------------------------------------
That is correct.

Explain how you were able to answer the question without access to the image - only the question_stem and choices.
What strategies did you use?

Then, briefly summarize these "language-only strategies" into a short list.

Finally, say whether your reasoning was closer to:
- 'deduction': you answered with high confidence because you had enough information.
- 'abduction': you answered with lower confidence you made an informed guess.

Be concise in your final response.

--------------------------------------------------------------------------------
Assistant
--------------------------------------------------------------------------------
To answer the question without the image, I relied entirely on analyzing the given information and the plausible characteristics of each option. Here's a breakdown of my strategy:

1. **Keyword and Context Analysis**: I focused on key descriptors such as "eye-ball" shaped features and their being "anomalous" in an electron microscopy image of neurons expressing human mHTT. This guided me to consider characteristics commonly seen under electron microscopy.

2. **Knowledge of the Subject Matter**: I applied general knowledge about electron microscopy artifacts and protein behavior. This helped me weigh the likelihood of each option based on what typically causes specific image features.

3. **Process of Elimination**: By understanding the nature of each option-including potential artifacts or structures (such as vesicles or ice) in microscopy-I systematically assessed the validity of each scenario in relation to the described features.

4. **Logical Inference**: I inferred which scenario most commonly results in anomalies resembling "eye-ball" features, favoring the simplest explanation.

Summary of "language only strategies":
- Analyze keywords and context
- Apply relevant subject matter knowledge
- Use process of elimination
- Make logical inferences

The reasoning was closer to 'abduction' as I made an informed guess based on patterns and typical causes, given that I lacked direct visual confirmation.
\end{lstlisting}

\item Next, we call \verb|rewrite_question|. Here, the templated variable \verb|{{conversations}}| has the full conversations produced by the function \verb|analyze_why_answerable| in the previous call and from all other iterations. The \verb|{{n_chat}}| just indicates how many conversations there are included:
\begin{lstlisting}[
    basicstyle=\ttfamily\scriptsize,
    breaklines=true,
    columns=flexible,
    frame=single,
    backgroundcolor=\color{prompt_bg},
    showstringspaces=false,
    mathescape=true,
    xleftmargin=2em,
    framexleftmargin=1.5em
]
Below, I will display {{n_chat}} chat conversations between a 'user' and an LLM 'assistant'.

In each conversation
	- a user asks the assistant to answer a multichoice VQA question, however they do not provide the image. They only get the question_stem and choices.
	- the question_stem and correct answer are similar to other conversations, however the distractors are different
	- the assistant then answers correctly.
	- the user then asks the assistant to explain how it answered the question with only the text.
	- the assistant then summarizes what strategy they used to answer the question. 
Altogether, these conversations give examples of language shortcuts that need to be avoided when constructing the question_stem and choices.

Here are the conversations:

{{conversations}}

Your task is to rewrite the question_stem and choices so that a different LLM 'assistant' cannot use the language-only strategies that were identified in these past conversations.
Include an 'explanation' about why your new set of distractors are better.
Your revised choices should include the correct answer at the 'correct_index'.

Your revised question_stem and choices should not significantly change the meaning of the question and correct answer.
You are free to change the distractors a lot to achieve this task. 
Include {{n_choices}} choices.
The choices should not be made too different in length and detail to the original question stem and choices.
Note that the question stem may contain important cues that cannot be removed. E.g. if a question asks about the "green stain" in an image, you cannot change it to "stain" because it introduces ambiguity.
The choices should not include the letters in front like (a), (b), ...

Return a json
\end{lstlisting}

The output of this function is a JSON and is:
\begin{lstlisting}[
    basicstyle=\ttfamily\scriptsize,
    breaklines=true,
    columns=flexible,
    frame=single,
    backgroundcolor=\color{prompt_bg},
    showstringspaces=false,
    mathescape=true,
    xleftmargin=2em,
    framexleftmargin=1.5em
]
{
  "question_stem": "In an electron microscopy image of a neuron expressing human mHTT (BACHD) protein, distinct, spherical features are prominent and seem to be atypical. What is the most probable explanation for these features?",
  "choices": [
    "Concentration of cellular waste products leading to distinct morphological anomolies",
    "Formation of air bubbles introduced during sample immobilization processes",
    "Ice contamination from liquid nitrogen during sample preparation",
    "Inclusion bodies formed by aggregated proteins expressing during stress response",
    "Cellular morphological changes during mitotic division"
  ],
  "correct_index": 2,
  "explanation": "The revised options introduce plausible scientific phenomena that could also result in spherical or distinct anomalies. This set of distractors is more varied and grounded in legitimate alternative explanations without leading the responder towards the correct answer through language patterns. For example, air bubbles and protein inclusion bodies might mimic sphere-like structures. Additionally, by adding an unrelated phenomenon like cellular changes during division, the set confuses the language pattern strategy previously used. Thus, the correct choice of \"Ice contamination from liquid nitrogen\" stands out less and necessitates either visual confirmation or real understanding of sample preparation outcomes to identify correctly."
}
\end{lstlisting}

Notice that the response includes an explanation of why it made the changes, and the changes relate to what was in the reflection, adding interpretability.

\item Finally, we check that the rewrite has not changed the semantic meaning of the question and answer. 
\begin{lstlisting}[
    basicstyle=\ttfamily\scriptsize,
    breaklines=true,
    columns=flexible,
    frame=single,
    backgroundcolor=\color{prompt_bg},
    showstringspaces=false,
    mathescape=true,
    xleftmargin=2em,
    framexleftmargin=1.5em
]
Prompt
********************************************************************************
Below are two question-answer pairs.
The question-answer pairs are part of VQA triplets, and both pairs use the same image.

Are these question-answer pairs semantically equivalent?
Or are they significantly different?
Give a true/false and also an explanation.

QUESTION 1:
In an electron microscopy image of a mouse primary embryonic cortical neuron expressing full-length human mHTT (BACHD), "eye-ball" shaped features are observed in the top left corner that appear anomalous. What is the most likely cause of these features?

ANSWER 1:
Ice contamination from liquid nitrogen during sample preparation

QUESTION 2:
In an electron microscopy image of a neuron expressing human mHTT (BACHD) protein, distinct, spherical features are prominent and seem to be atypical. What is the most probable explanation for these features?

ANSWER 2:
Ice contamination from liquid nitrogen during sample preparation

********************************************************************************
Response, is_equivalent: True
********************************************************************************
is_equivalent: Both question-answer pairs revolve around electron microscopy images depicting neurons expressing the human mHTT (BACHD) protein. The main focus in each question is on the presence of unusual spherical (or 'eye-ball' shaped) features. In both cases, the identical explanation is provided: ice contamination from liquid nitrogen during sample preparation.

While the wording differs slightly, the core components, contexts, and the ultimate explanation remain consistent between the two pairs. Thus, they can be considered semantically equivalent as they describe the same phenomenon with the same reasoning for the observed anomalies.
\end{lstlisting}

\item If the meaning check step fails, then we exit with a fail as we notice that once the meaning deviates it will become more distorted with further iterations. Otherwise, we update the MCQ, and start the loop again. Where the first function call in the new loop will check whether this revised MCQ is now too hard or needs to be rewritten.
\end{enumerate}

\paragraph{Calibrating the preservation of meaning}
Since there is a risk that the rewriter will change the semantic meaning of the question and answer, it is crucial that the \verb|check_meaning_preserved| is effective. For topics requiring domain expertise, this requires prompt engineering with a domain expert providing feedback. This is one of the key human bottlenecks in tuning the prompts in this system.

\paragraph{Other sources of MCQ invalidation}
While the \verb|check_meaning_preserved| function checks that the question and answer is correct, an additional risk is that the rewrite introduces distractors that are more correct than the target answer. In our manual review evaluations (we reviewed all 1,042 MCQs), this was very rare. However, if this were a significant problem in future systems, it may be addressable by tuning the rewriter bot, with an expert human evaluator in the prompt engineering loop.

\paragraph{Extending RefineBot beyond language shortcuts}
We constructed RefineBot specifically to address the issue of language shortcuts -- where the question is answerable without using the image. This is appropriate because we work with vision-centric VQA, meaning the questions should require image interpretation. However, the framework should work in other settings -- namely language-only QA. This would require adjusting the prompts slightly. For example, the current reflection prompt asks how the question was successfully evaluated without using language; also, the current rewriter prompt directs the LLM to address language shortcuts in its rewrite. Therefore, adapting Refinebot to language-only MCQ should make these prompts more general.

On the other hand, the method could also be updated for the more general VQA setting -- to improve questions that do not have language shortcuts, but are easy for some other reason. Here, the evaluator and reflector agents, which are currently LLMs, could be switched for an MLLM, which also ingests the image (though this would be expensive for current state-of-the-art models). We hypothesize that current MLLMs are so weak at image perception, that this is not necessary for  MicroVQA -- that simply removing language shortcuts will suffice for creating a challenging benchmark. As MLLMs improve at perceiving microscopy images, this general VQA RefineBot could become more useful for ensuring that the benchmark remains challenging. 

\paragraph{Repeated RefineBot iterations}
A single run of RefineBot may exit as a FAIL, either because the rewrite changed the meaning of the question, or because the maximum number of iterations was reached. However, RefineBot can give very different outputs when changing the random seed, and we find that simply running again with a different random seed can lead to a successful run. Therefore, if the bot fails, we rerun it.

If an MCQ exited RefineBot successfully, then the evaluator LLM was fooled by the question (got the answer incorrect). Yet, the question could still be suboptimal in detectable ways: the LLM could be capable of narrowing the MCQ down to two options, but have guessed the wrong option; moreover, this question could be easier for different models. To ensure it is difficult enough overall, we require that the MCQ is answered incorrectly for two random seeds from two different models: \verb|gpt-4o-2024-08-06| and \verb|claude-3-5-sonnet-20241022|. If not, then it fails, and we run the RefineBot again with a different random seed. 

We discuss in the experiments \cref{fig:results-main} how this process leads to a small evaluation bias against the models used as evaluators. That section shows that despite the bias, this process makes the MCQs generally harder for all models. Future versions of RefineBot could mitigate this by including a larger family of evaluator models. 

Finally, some MCQs fail to pass the RefineBot check. In these cases, we simply keep the stage 1 MCQ.


\subsection{MCQ generation results: stage 1 exam alignment}
The Stage 1 pipeline produced one-best-answer MCQs that preserved the content, testing focus, and validity of the expert-generated question-answer pairs. It generated an initial set of distractors following best practices outlined in educational and NBME guidelines. A key design goal of Stage 1 was to preserve the original question content without altering or introducing new information. Secondary objectives included ensuring proper formatting according to NBME guidelines and minimizing information that might provide clues to the correct answer without reference to the image.

While the pipeline ensured alignment with these goals, preserving the original content occasionally resulted in MCQs containing extra information, such as image descriptions or specific biological terms (i.e., gene or protein name), that could inadvertently help narrow the options. For example, stage 1 question stems sometimes included protein names, disease descriptions, or descriptions of image features, which could reduce the challenge. Figure \ref{fig:qual_stage2} illustrates Stage 1 outputs that are appropriately formatted but may preserve details from the original input, making questions less challenging. While the distractors were biologically plausible, some options were overly easy to eliminate, reducing the challenge. These limitations are addressed and refined in Stage 2.

The \texttt{o1-mini-2024-09-12} LLM judge, employed during DSPy optimization, compared the Stage 1 MCQs to the original inputs to evaluate (1) semantic similarity and (2) adherence to MCQ formatting. The results showed that 96\% of questions preserved the original meaning, and 100\% adhered to NBME formatting guidelines. Manual analysis of the 4\% labeled as at least partially dissimilar revealed that these cases typically involved open-ended original questions being reformulated into closed-ended MCQs. This transformation, a necessary step for creating one-best-answer MCQs, was not a failure but rather an expected outcome of Stage 1. In rare instances where the original submission included multiple correct answers, the output of Stage 1 retained one correct answer and excluded others to comply with the one-best-answer format. The LLM judge flagged these cases as partially dissimilar but these changes were essential for adhering to the MCQ structure.

A common issue in MCQ design is that correct answers tend to be longer and more detailed than distractors. To assess this, we computed the ratio of tokens in incorrect options to those in the correct answer, where values $\leq 1$ indicate longer correct answers, and values $\geq 1$ suggest shorter correct answers. Stage 1 MCQs showed a ratio of 0.87, indicating that correct answers were only slightly longer than distractors. By comparison, naive MCQ generation yielded a ratio of 0.46, meaning correct answers were nearly twice as long as distractors, making them easier to identify. Additionally, naive MCQ generation preserved the original meaning in only 89\% of cases and met NBME formatting guidelines in just 70\%, highlighting the efficacy of the Stage 1 process at preserving information and ensuring proper MCQ formatting.

\subsection{MCQ generation results: stage 2 RefineBot}
\label{sec:appendix-stage2-analysis}
\begin{figure}[ht]
    \centering
    \includegraphics[width=0.8\linewidth]{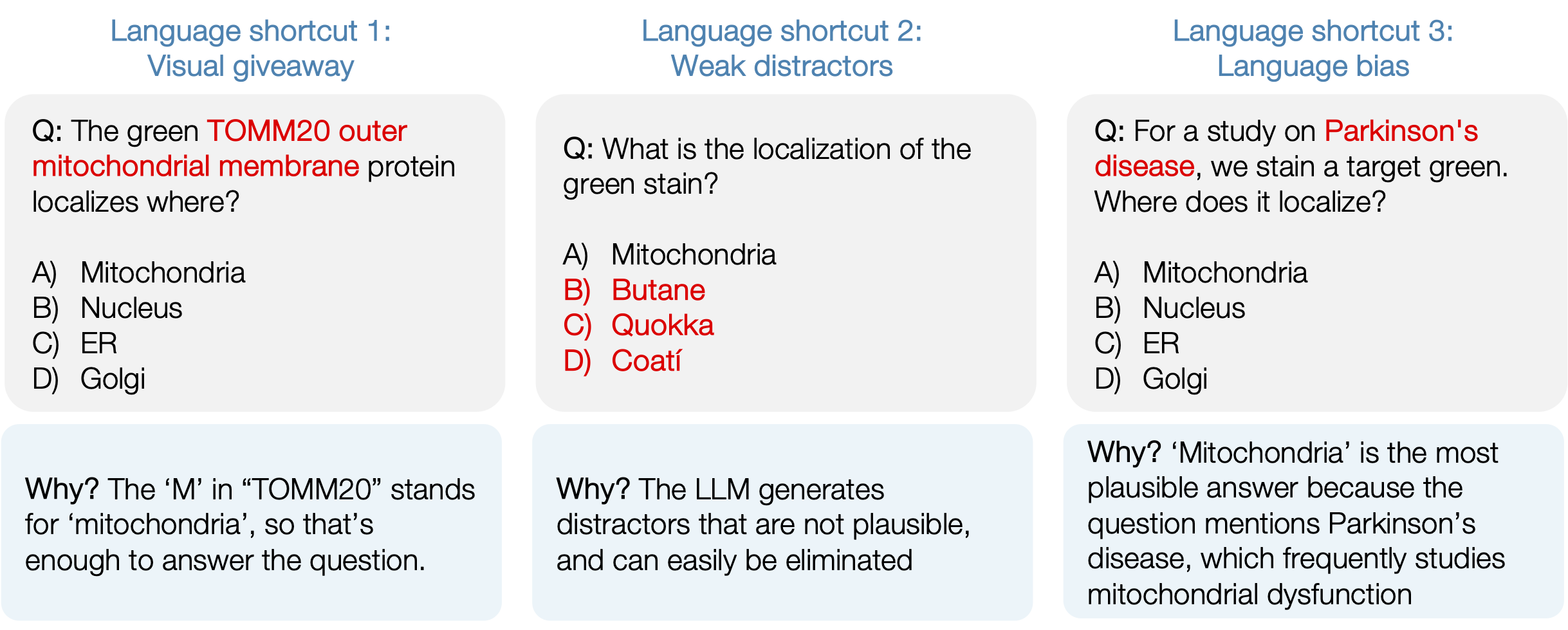}
    \caption{Three types of language shortcut relevant to MicroVQA. The target VQA has an image that is fluorescence microscopy stained with TOMM20 which would show a pattern consistent with visualizing mitochondria.}
    \label{fig:appendix-language-shortcuts-types}
\end{figure}

The second stage of our MCQ generation method focuses on increasing the difficulty of the questions and options. The key idea is that we want to remove language shortcuts -- information in the MCQ that allows answering the question without access to the image. To aid in understanding, we construct a toy example in \cref{fig:appendix-language-shortcuts-types} with the three types of language shortcuts that we identify. The questions should be paired with a stain that is mitochondria, and we want the VQA to require actually inferring that from the image.
In shortcut 1, the text `gives away' the image content so it's trivial to answer the question. In shortcut 2: The LLM generates implausible or weak distractors. In shortcut 3: 'Language bias', is a known problem in VQA \cite{goyal2017making}. 

\begin{figure}[h]
    \centering
    \includegraphics[width=\linewidth]{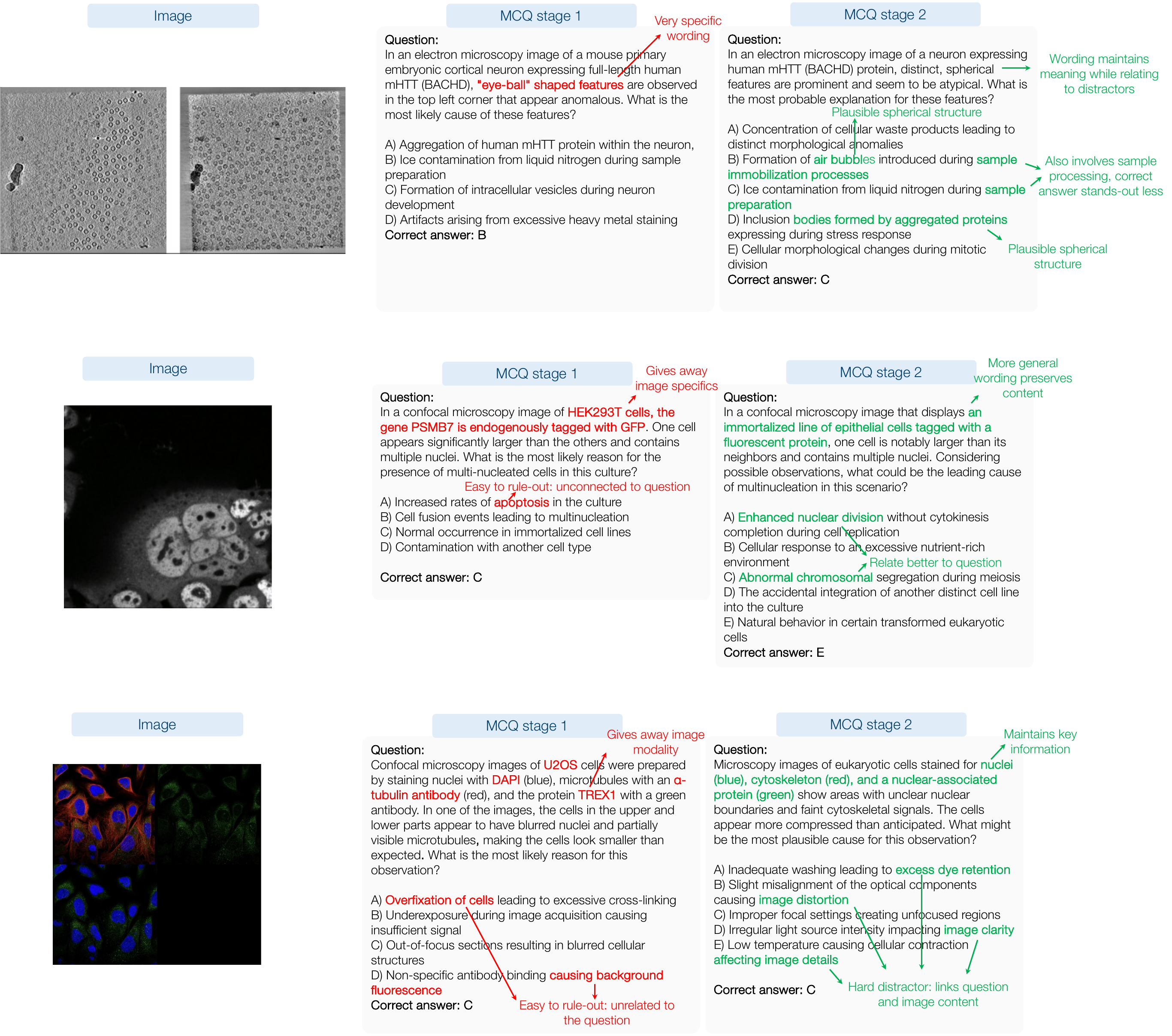}
    \caption{Examples of changes to questions and options between stage 1 and stage 2 (RefineBot) of our MCQ generation method. In \red{red} elements that need to be improved and in \color{Green} green \color{black} improvements.}
    \label{fig:qual_stage2}
\end{figure}

Moving to real MCQs, in \cref{fig:creation} we illustrate the transformation of a question at each stage, highlighting issues and corresponding improvements. While stage 1 produces exam-style questions that preserve the QA objective, the wording of the questions and distractors often enable MLLMs to exploit shortcuts rather than demonstrate knowledge. To address this, stage 2 (RefineBot) employs a chain-of-thought process. We analyze additional examples of improvements achieved in stage 2, showing how RefineBot improves both the style and content. The key strategies correspond to the three shortcut types that we identified:
\begin{enumerate}[leftmargin=1.5cm] 
    \item (Visual giveaway) Over-description of the image eliminates the need for image interpretation. For example, in \cref{fig:qual_stage2} (middle row), stage 1 includes details that reveal the specifics of the image, allowing the model to propose common issues without analyzing the image. Stage 2 replaces these specifics with general terms in the question stem.
    \item (Weak distractors) Distractors unrelated to the specifics of the scenario described by the question. In \cref{fig:qual_stage2} (bottom row), stage 2 increases distractor complexity by incorporating elements tied to the question's context, making them less predictable.
    \item (Language bias) Overly precise wording in the question stem: Precise phrasing can make it easy to identify the correct answer through pattern matching. For example, \cref{fig:qual_stage2} (top row), changing `eye-ball shaped' to `spherical' retains the question's intent while creating more challenging distractors referencing plausible alternative processes.
\end{enumerate} 

\FloatBarrier
\section{Experiments details}
\label{sec:appendix-results}

\subsection{Evaluation prompts}
All prompts except for LLaVA and LLaVA-Med used the chain-of-thought prompt template from the MMMU-Pro code \cite{yue2024mmmu_pro}:
\begin{mdframed}[backgroundcolor=prompt_bg, font=\small, leftmargin=0cm, rightmargin=0cm]
\scriptsize
\begin{lstlisting}
The following is a multiple choice question (with answers). 
Think step by step and then output the answer in the format of \"The answer is (X)\" at the end.

{{QUESTION}}

Options:
{{CHOICES}}
\end{lstlisting}
\end{mdframed}

An example complete question is:
\begin{mdframed}[backgroundcolor=prompt_bg, font=\small, leftmargin=0cm, rightmargin=0cm]
\scriptsize
\begin{lstlisting}
The following is a multiple choice question (with answers).
Think step by step and then output the answer in the format of "The answer is (X)" at the end.

A fluorescence microscopy image displays cultured mouse primary neurons stained with a specific marker localized to mitochondria. The fluorescence signal exhibits an uneven distribution across the neuronal cell, and there is an absence of signal within the nucleus. What is the most probable cause for the uneven distribution of the mitochondrial signal?

Options:
  (1): Mitochondrial transport along axonal and dendritic processes ensures uneven distribution based on cellular energy demands.
  (2): Differential expression of mitochondrial proteins leads to variable fluorescence signal across the neuron.
  (3): Mitochondrial biogenesis occurs at specific neuronal regions, causing localized fluorescence accumulation.
  (4): Mitochondrial anchoring to the cytoskeleton restricts their movement, resulting in uneven distribution.
  (5): Mitochondrial dynamics, including fission and fusion, allow mitochondria to move and distribute throughout the neuron.
\end{lstlisting}
\end{mdframed}

The answer is then extracted with the regex pattern:   \verb|answer is \(?([0-9])\)?|.

For LLaVA-Med, we added an extra parsing strategy. This was because we observed that while LLaVA-Med often gave reasonable responses to questions, it failed to follow instructions well. First, it did not follow the CoT prompt consistently to \verb|think step by step|. It also didn't use the right answer formatting: \verb|"The answer is (X)\"|. Since we were more interested in biomedical reasoning adhering to VQA formatting, we decided to handle this model differently.

Specifically, to encourage CoT reasoning, we fixed the prefix of the response to:\\
\noindent\hspace*{1cm}\verb|Response: Let's think through this step by step and evaluate each option.|.  \\
\noindent Then we added a suffix to the end:\\
\noindent\hspace*{1cm}\verb|After carefully considering all the options, the most likely answer is (|, \\
\noindent Then we continued generation, and did regular regex parsing to get the final answer.

We also find that a few models make minor but consistent errors with formatting the answer and elect to fix them. For example \verb|x-ai/grok-2-vision-1212| commonly adds two asterisks before the answer like ``The answer is **(4)**'' instead of ``The answer is (4)'' (\verb|GPT-4o| also did this sometimes). Some other common errors that we saw but did not handle (because it would be difficult to do rigorously) was shown by \verb|meta-llama/llama-3.2-11b-vision-instruct|: the response would write one of the MCQ options but not state its number.

\subsection{Model details}
Our main benchmarking experiments in \cref{tab:results} refer to models that have multiple versions. The exact model versions used for inference are shown in \cref{tab:model-endpoints}
\begin{table}[htbp]
\footnotesize
\centering
\begin{tabular}{lll}
\hline
\textbf{Model Name} & \textbf{API Endpoint} & \textbf{Source \& Details} \\
\hline
o1 & \texttt{o1-2024-12-17} & OpenAI API \\
Claude-3.5-Sonnet & \texttt{*anthropic/claude-3.5-sonnet-20240620} & Openrouter API \\
Gemini-Pro-1.5 & \texttt{google/gemini-pro-1.5} & Openrouter API \\
Pixtral-Large & \texttt{mistralai/pixtral-large-2411} & Openrouter API \\
Grok-2-Vision & \texttt{x-ai/grok-2-vision-1212} & Openrouter API \\
Qwen-2-vl-72b-Instruct & \texttt{qwen/qwen-2-vl-72b-instruct} & Openrouter API \\
VILA1.5-40b & \texttt{VILA1.5-40b} & HuggingFace, local inference \\
GPT-4o & \texttt{gpt-4o-2024-08-06} & OpenAI API \\
Llama-3.1-Nemotron-70b-Instruct & \texttt{nvidia/llama-3.1-nemotron-70b-instruct} & Openrouter API \\
Llama-3.2-90b-Vision-Instruct & \texttt{meta-llama/llama-3.2-90b-vision-instruct} & Openrouter API \\
Qwen-2-VL-7b & \texttt{qwen/qwen-2-vl-7b-instruct} & Openrouter API \\
Claude-3.5-Haiku & \texttt{anthropic/claude-3.5-haiku} & Openrouter API \\
Gemini-Flash-1.5-8b & \texttt{google/gemini-flash-1.5-8b} & Openrouter API \\
GPT-4o-mini & \texttt{gpt-4o-mini-2024-07-18} & OpenAI API \\
Pixtral-12b & \texttt{mistralai/pixtral-12b} & Openrouter API \\
VILA1.5-13b & \texttt{VILA1.5-13b} & HuggingFace, local inference \\
Llama-3.2-11b-vision-instruct & \texttt{meta-llama/llama-3.2-11b-vision-instruct} & Openrouter API \\
LLaVA-Med-Mistral-7B & \texttt{LLaVA-Med-Mistral-7B} & HuggingFace, local inference \\
LLaVA-Mistral-7B & \texttt{LLaVA-Mistral-7B} & HuggingFace, local inference\\
\hline
\end{tabular}
\caption{AI model API endpoints and their sources}
\label{tab:model-endpoints}
\end{table}

For LLaVA and LLaVA-med, although the version number 1.6 is different from LLaVA-Med with 1.5, we do believe that LLaVA-Med was either trained from this LLaVA version, or from one that is very similar. They have identical architecture, and the version is very close; it is possible the LLaVA-Med copied the number incorrectly. The LLaVA-Med codebase or github issues does not clarify this at the time of writing. 

Note that for OpenRouter function calls to Anthropic, the date is not included in the string. We believe they use timestamps for older model versions, but not for the latest version. So reproducing it may require checking if the model has been updated.

\subsection{Human baseline on MicroVQA}
\label{sec:results-humanbaseline}
To perform a human baseline, we recruited co-authors who created questions. The conditions were:

\begin{itemize}
    \item Questions were provided in a google form.
    \item Participants did not view any question that they created. 
    \item Internet access was allowed, but LLM access was not. This is because realistic biological practice is knowledge-intensive, and may require looking up facts or references.
    \item Time limit was 3 minutes per question. They were free to take as many breaks as they wanted. 
    \item They answered the multiple choice question without further written justification.
\end{itemize}

These conditions -- 3 minutes and internet access -- give sufficient time for an expert to make a reasonable attempt. Since the score was far below 100\%, we claim that these questions could be of value to real experts as an LLM chat assistant. Each participant only took a subset of the questions (about 100 each). The final score is the aggregate.

\subsection{More experiments on attribute-level information}
\begin{figure}[h]
    \centering
    \includegraphics[width=0.35\linewidth]{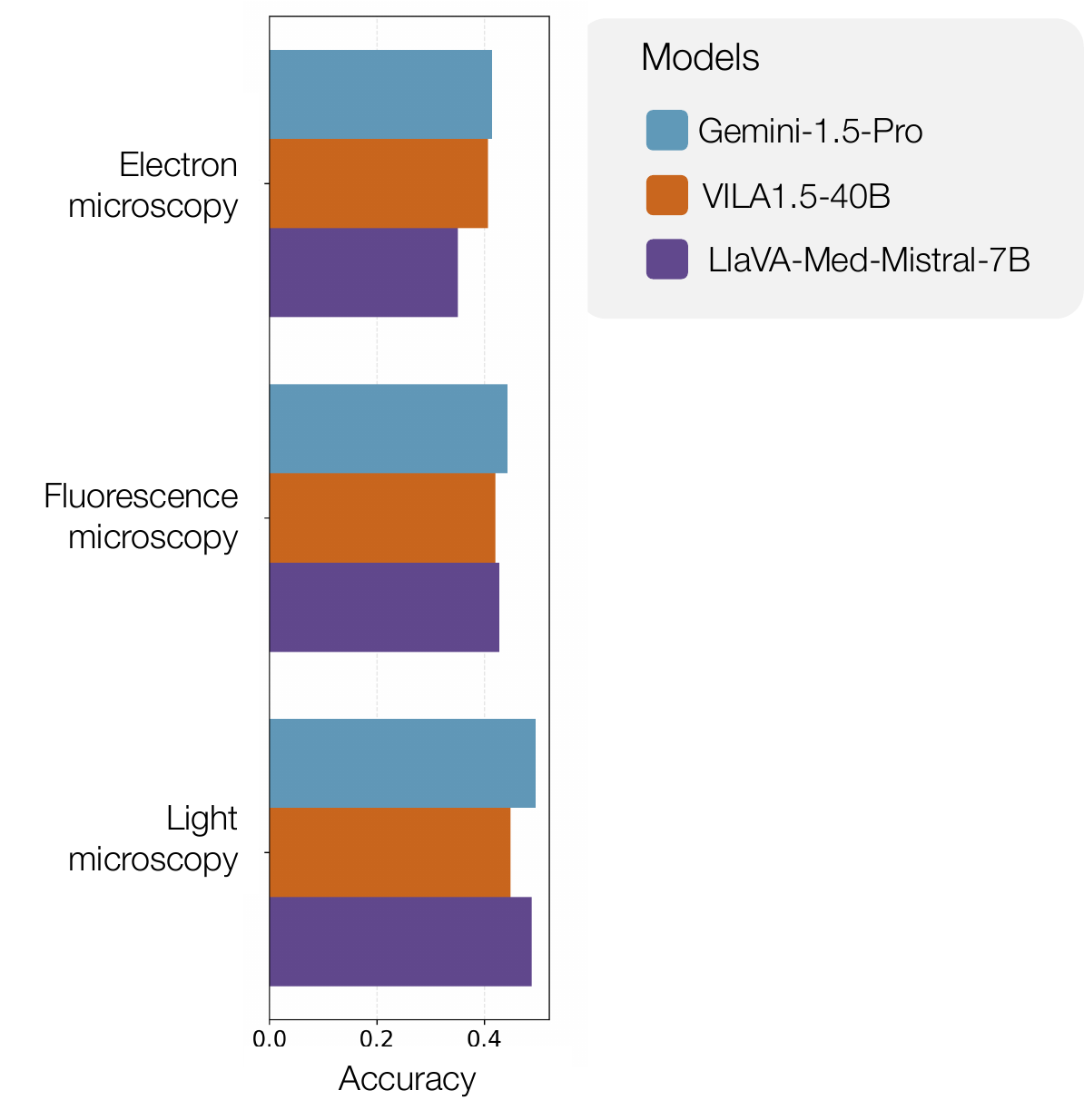}
    \caption{Performance by image modality type for the best models: Gemini-1.5-Pro (closed source), VILA1.5-40B (open-source), and LlaVA-Med-Mistral-7B (medical).}
    \label{fig:appendix-modality}
\end{figure}
We described our metadata features in \cref{sec:benchmark}, and we can now leverage them to benchmark models on different attributes.  \cref{fig:analysis_modality} (left) shows all sub-tasks, and shows that `compare image sets', which is common in biology to compare `treatment' and `control' groups. All models perform relatively well, despite multi-image reasoning being understudied in MLLMs \cite{wang2024muirbench}. \cref{fig:analysis_modality} (right) shows that for Gemini and LLaVa-Med, higher Bloom's levels -- corresponding to more advanced reasoning -- leads to worse performance. This supports the intuitive result that higher-level reasoning is more challenging -- although VILA does not follow this trend. A final result in \cref{fig:appendix-modality} shows that all models performed worse on cryo-electron microscopy images than brightfield or fluorescence microscopy. This may be due to cryo-EM's challenging features (low contrast, low signal-to-noise ratio) and their under-representation in pretraining datasets \cite{zhang2023biomedclip}.

\subsection{Language shortcut ablations} 
\label{sec:appendix-language-ablations}
In \cref{tab:mcq_ablation} we do language shortcut ablations for some top-performing models and some lower-performing models -- both no-image and choices-only ablations. 

The no-image ablation is a common test in VQA evaluations\cite{tong2024cambrian, goyal2017making}. The MLLM only receives the text prompt, and we add the following sentence to the prompt template:\\
\noindent\hspace*{1cm}\texttt{If an image is mentioned ignore this information and try your best to answer the question.}\\
The choices-only ablation has been explored in (non-visual) question answering \cite{balepur2024artifacts}. The model does not receive the image or the question and must guess an answer from only the choices. We add this sentence to the prompt template:\\
\noindent\hspace*{1cm}\texttt{However I will not give you the question text or the images, I will only give you the choices, so please try your best to answer the question.}

\begin{table}[t]
\caption{Ablation study on MicroVQA MCQs. Column 2 is overall accuracy. Cols 3-4 are the no-image ablation accuracy and drop compared to overall accuracy. Cols 5-6 are the choices-only ablation accuracy and drop compared to overall accuracy. See the text for discussion.}
\centering
\begin{tabular}{@{}lc|cc|cc@{}}
\toprule
 & \multicolumn{1}{l|}{} & \multicolumn{2}{l|}{No-image ablation} & \multicolumn{2}{l}{Choices-only ablation} \\
 & \multicolumn{1}{l|}{Accuracy} & \multicolumn{1}{l}{Accuracy} & \multicolumn{1}{l|}{Drop} & \multicolumn{1}{l}{Accuracy} & \multicolumn{1}{l}{diff} \\
 \hline
o1 & 52.8 & 49.2 & -3.6 & 37.7 & -15.1 \\
Claude-3.5-Sonnet & 51.7 & 46.0 & -5.8 & 44.0 & -7.7 \\
Gemini-Pro-1.5 & 51.1 & 47.2 & -3.8 & 36.8 & -14.3 \\
Pixtral-Large & 49.8 & 46.3 & -3.6 & 36.7 & -13.1 \\
Grok-2-Vision & 48.4 & 46.3 & -2.1 & 40.5 & -7.9 \\
GPT-4o-mini & 46.2 & 44.2 & -1.9 & 34.3 & -11.9 \\
Pixtral-12b & 45.6 & 43.7 & -1.9 & 31.8 & -13.8\\
\hline
Random & 22.0 &  &  &  & \\
\bottomrule
\end{tabular}
\end{table}

Both ablations lower the scores, but they are still above random. While this is initially surprising, it is actually consistent with other benchmarks. A key challenge in VQA benchmarks is for questions to be ``vision-centric'': that the question requires using the image features to answer. We still have very strong evidence that MicroVQA is vision-centric, which we discuss below.  

First, let's recall the language shortcuts by considering a VQA sample with an image of a green field, with the question ``what color is the grass''.
\begin{itemize}
    \item `Language bias' is well-known for enabling models to make educated guesses about MCQs without access to the image \cite{goyal2017making}. For the question ``what color is the grass?'', then without the image, ``green'' is a likely guess, although there are other possible answers, like ``brown''.
    \item `Visual giveaway' means there are too many image details in the question, for example if the question were  ``what color is the green grass?'' This removes any need to look at the image.
    \item `Weak distractors' gives other question choices that are easily eliminated. For example, for question ``what color is the grass?'', a bad distractors would be ``pink''. This also removes any need to look at the image.
\end{itemize}
(In \cref{sec:appendix-stage2-analysis}, we also discussed language shortcuts with biology examples as it related to MCQ construction.)
Now we argue that MicroVQA is still vision-centric despite the ablation results.

\textbf{1. Language bias allows cheating on vision-centric questions} If answerable due to a `language bias', the question may still require perception. A model may guess that the grass is green, but it is only a guess -- the grass really could be brown \cite{goyal2017making}. Therefore correct answers under the `no-image' ablation may still be vision-centric. On the other hand problems with `visual giveaway' and `weak distractor' shortcuts really do render the question not-vision-centric. The RefineBot method we employed in MCQ generation was designed to target language shortcut removal. Anecdotally, our qualitative analysis (from \cref{sec:appendix-stage2-analysis}) found that RefineBot tends to focus on fixing the visual giveaway and weak distractor shortcuts. Ideally we would perform a quantitative analysis to identify which language shortcuts are due to language bias versus the others, however analyzing these questions is very time-consuming. 

\textbf{2. The unsolved questions are hard because of perception}. The `no-image' ablation only tells us about the `easy' questions, but half the quesitons are too hard for current models. Our qualitative analysis of 30 failure cases demonstrates that half of them made critical perception errors, which strongly suggests they require image understanding

\textbf{In contrast to the `vision-centric' claims from the Cambrian paper \cite{tong2024cambrian}}: They show that many VQA benchmarks have no-image ablation scores only slightly underperforming the with-image scores, and they conclude that such VQA benchmarks ``do not properly measure vision-centric capabilities''. But that conclusion cannot be drawn from only the no-image ablation (following our point 2 above). For benchmarks where MLLM's score far from 100\%, the failures could mostly be due to poor image understanding. This is very plausible since visual capabilities in MLLMs are viewed as lagging behind language capabilities. Likewise, our experience with our microscopy data was that visual understanding was poor, but that language knowledge was impressive. 

In point 1 above, we argued that  MLLMs can use language biases to pass the no-image ablation, even if they are vision-centric questions. This is actually in agreement with the claim from Cambrian that those questions don't test vision-centric capabilities. 

\textbf{Finally, some non-vision centric questions are acceptable}. The goal of MicroVQA is to source realistic problems from microscopy research. While this area does lean heavily on image analysis, not every question will realistically require perception, so this is acceptable.

\newpage
\subsection{Error analysis} \label{sec:appendix-error-analysis}
Here we show samples from the manual qualitative error analysis (details in the main paper). To see all 30 examples, refer \href{https://github.com/jmhb0/microvqa}{the github repo}.
\subsubsection{Perception errors}
\label{sec:appendix-error-analysis-perception}

\begin{figure}[h]
    \centering
    \includegraphics[width=\linewidth]{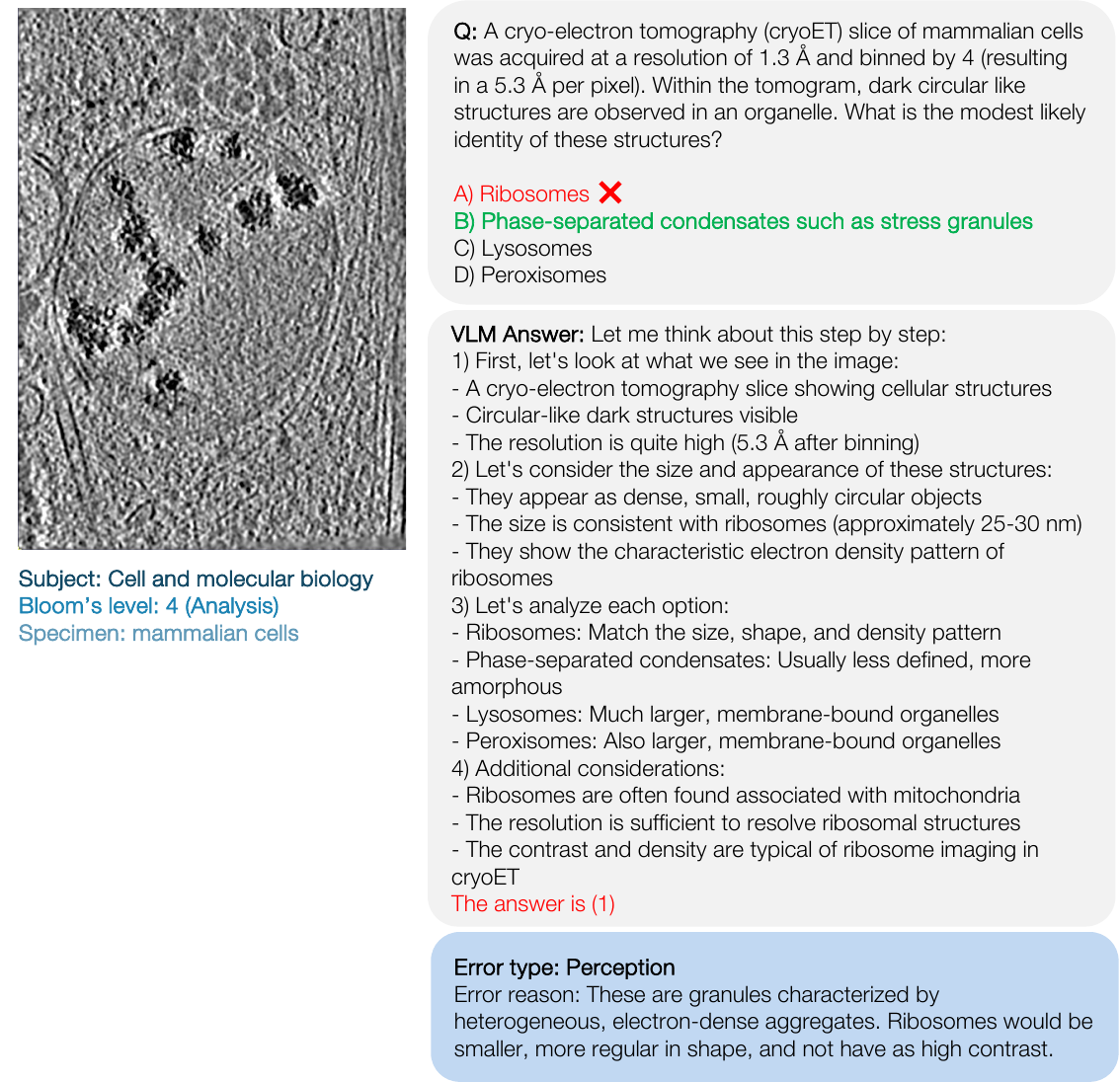}
    \caption{}
    \label{fig:error_perception_00}
\end{figure}

\newpage
\begin{figure}[h]
    \centering
    \includegraphics[width=\linewidth]{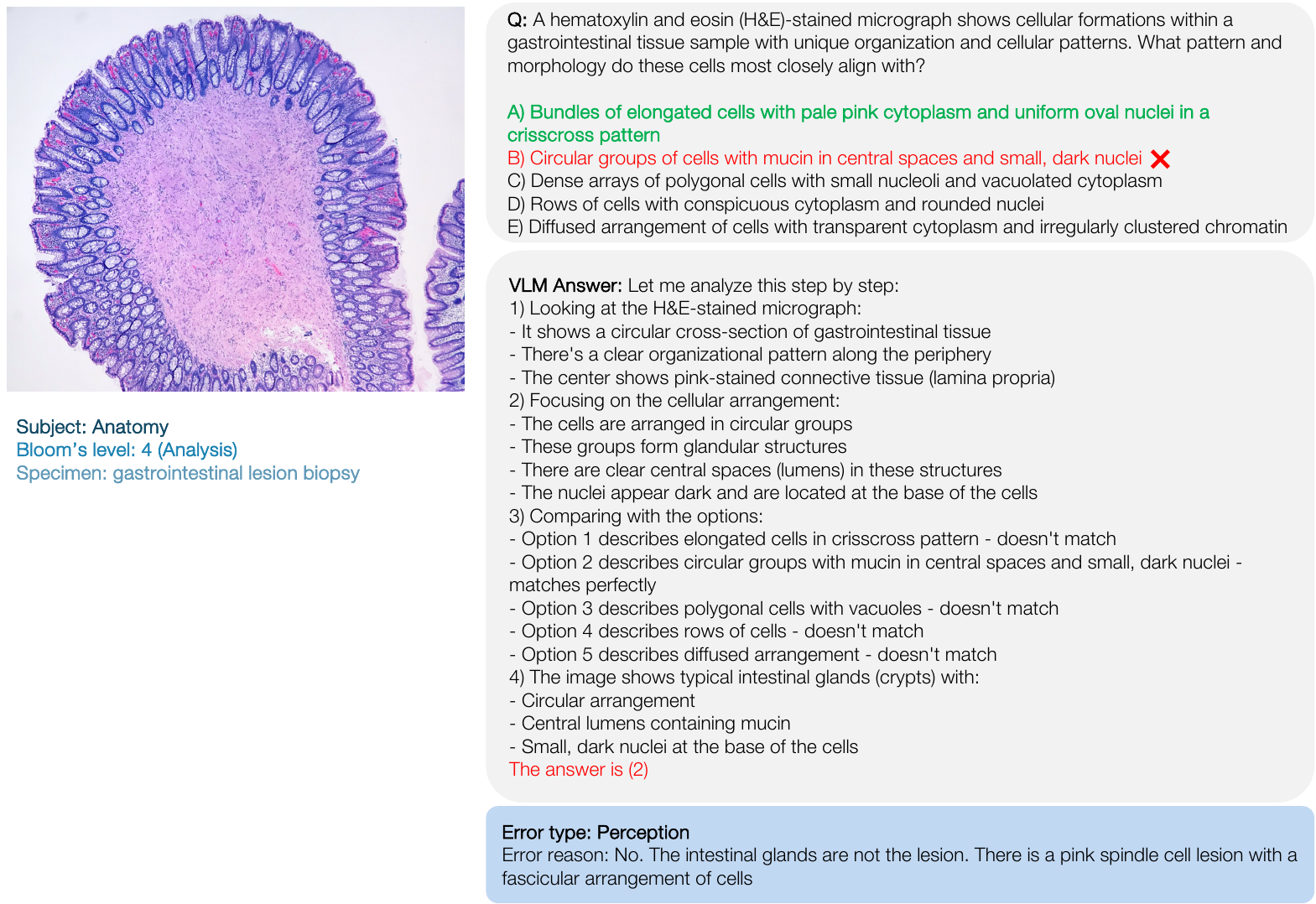}
    \caption{}
    \label{fig:error_perception_01}
\end{figure}

\newpage
\begin{figure}[h]
    \centering
    \includegraphics[width=\linewidth]{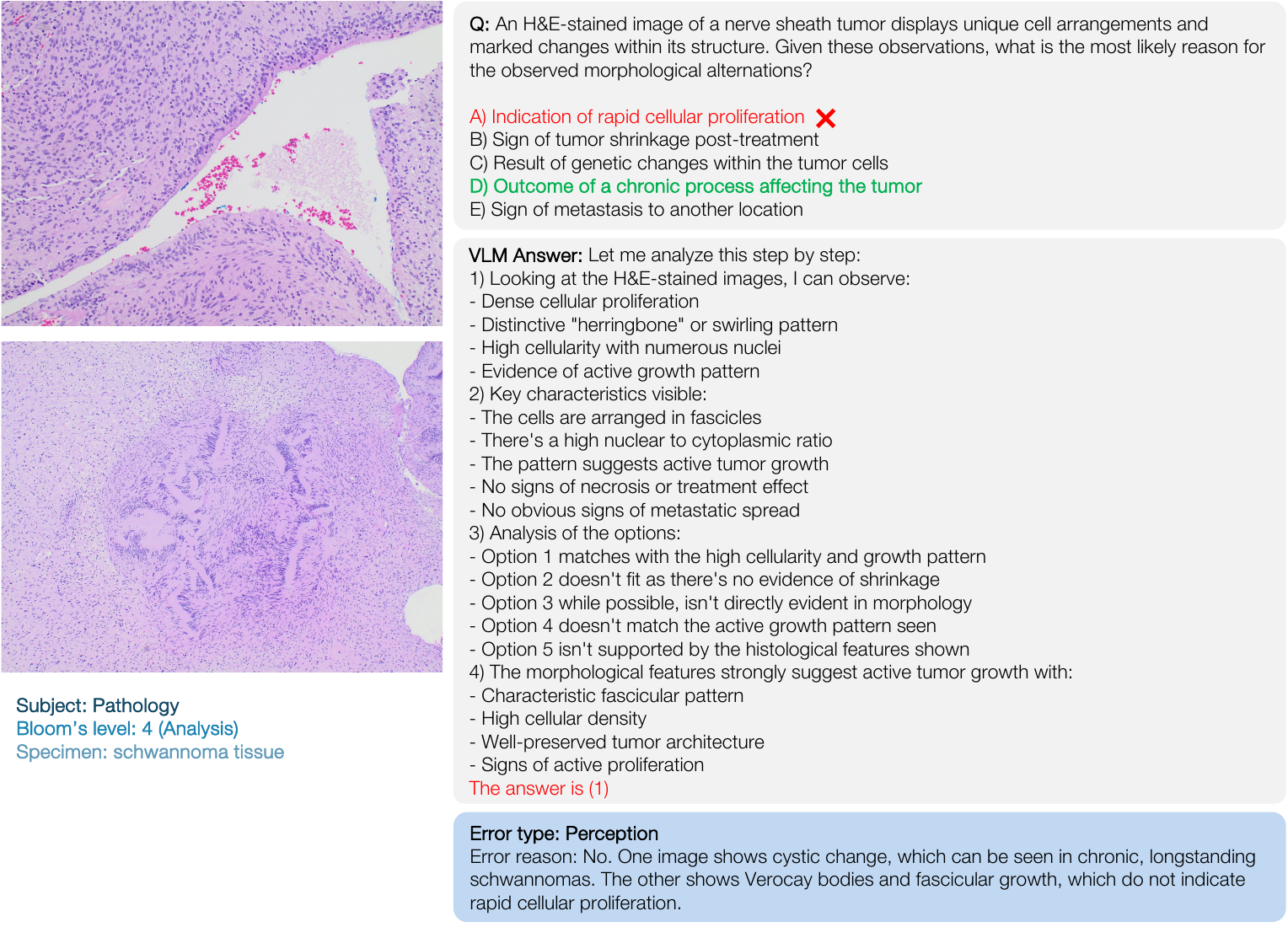}
    \caption{}
    \label{fig:error_perception_02}
\end{figure}

\newpage
\begin{figure}[h]
    \centering
    \includegraphics[width=\linewidth]{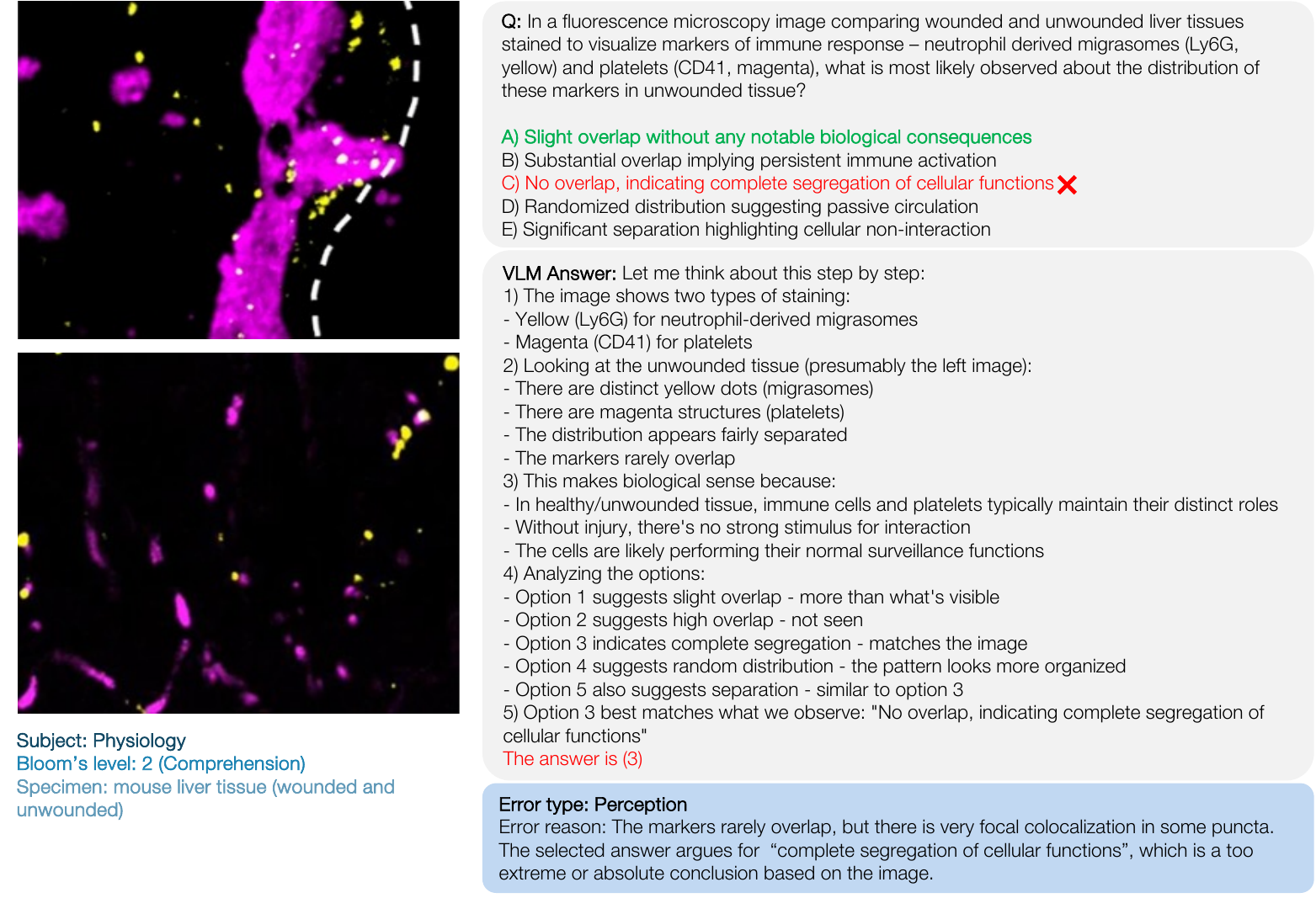}
    \caption{}
    \label{fig:error_perception_03}
\end{figure}



\newpage
\subsubsection{Knowledge errors}
\label{sec:appendix-error-analysis-knowledge}
\begin{figure}[h]
    \centering
    \includegraphics[width=\linewidth]{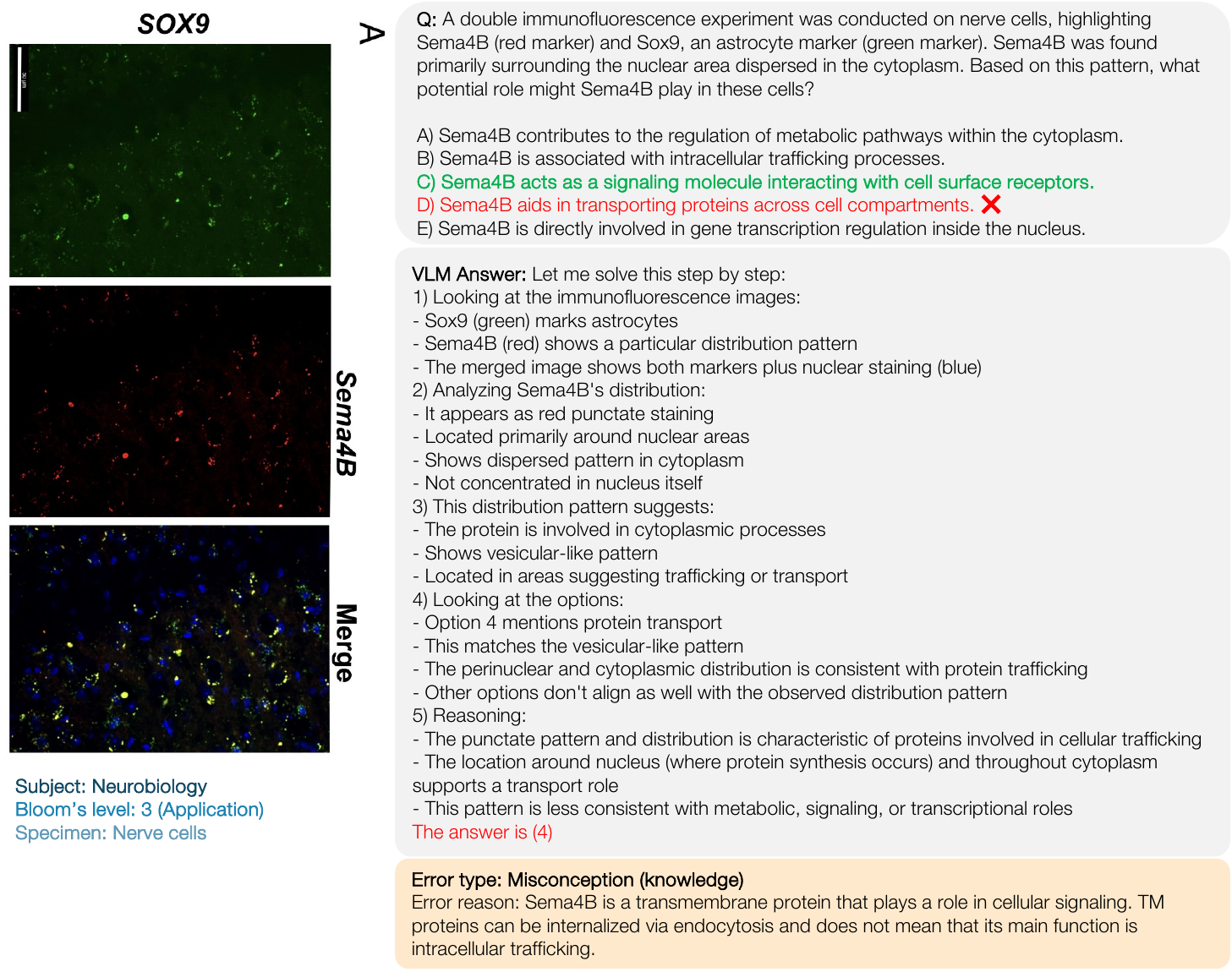}
    \caption{}
    \label{fig:error_knowledge_01}
\end{figure}

\newpage
\begin{figure}[h]
    \centering
    \includegraphics[width=\linewidth]{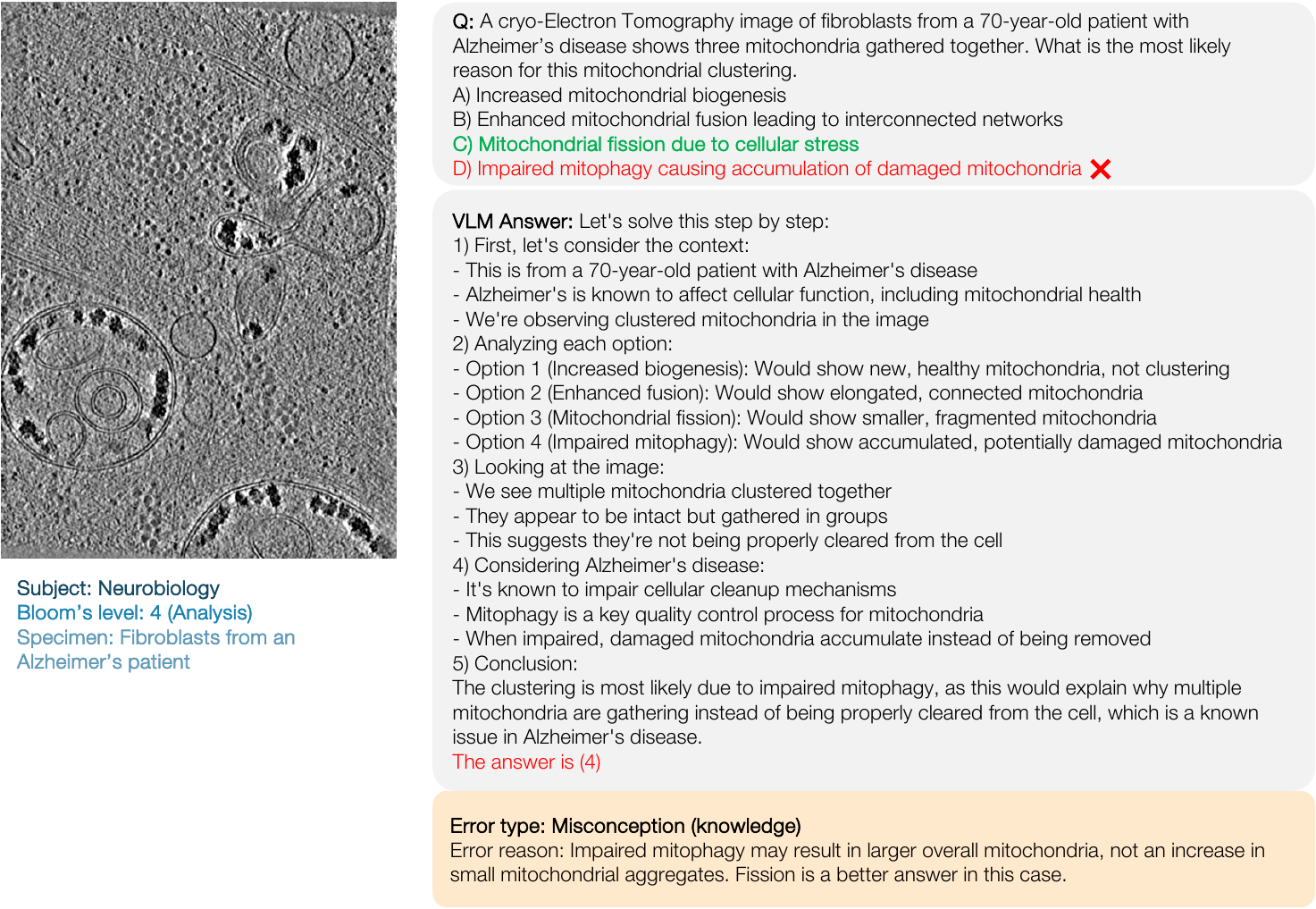}
    \caption{}
    \label{fig:error_knowledge_00}
\end{figure}

\newpage
\begin{figure}[h]
    \centering
    \includegraphics[width=\linewidth]{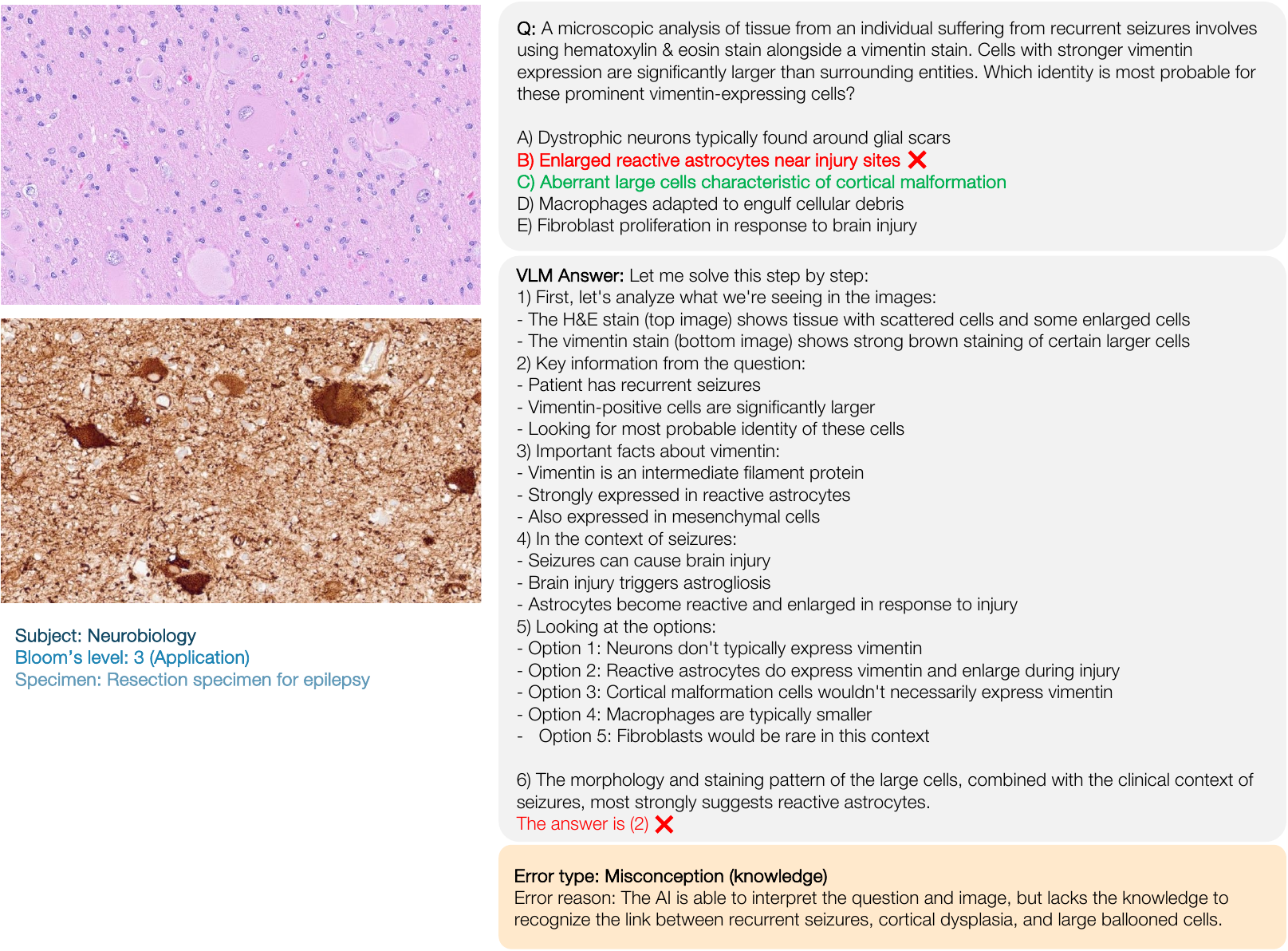}
    \caption{}
    \label{fig:error_knowledge_01_}
\end{figure}

\newpage
\begin{figure}[h]
    \centering
    \includegraphics[width=\linewidth]{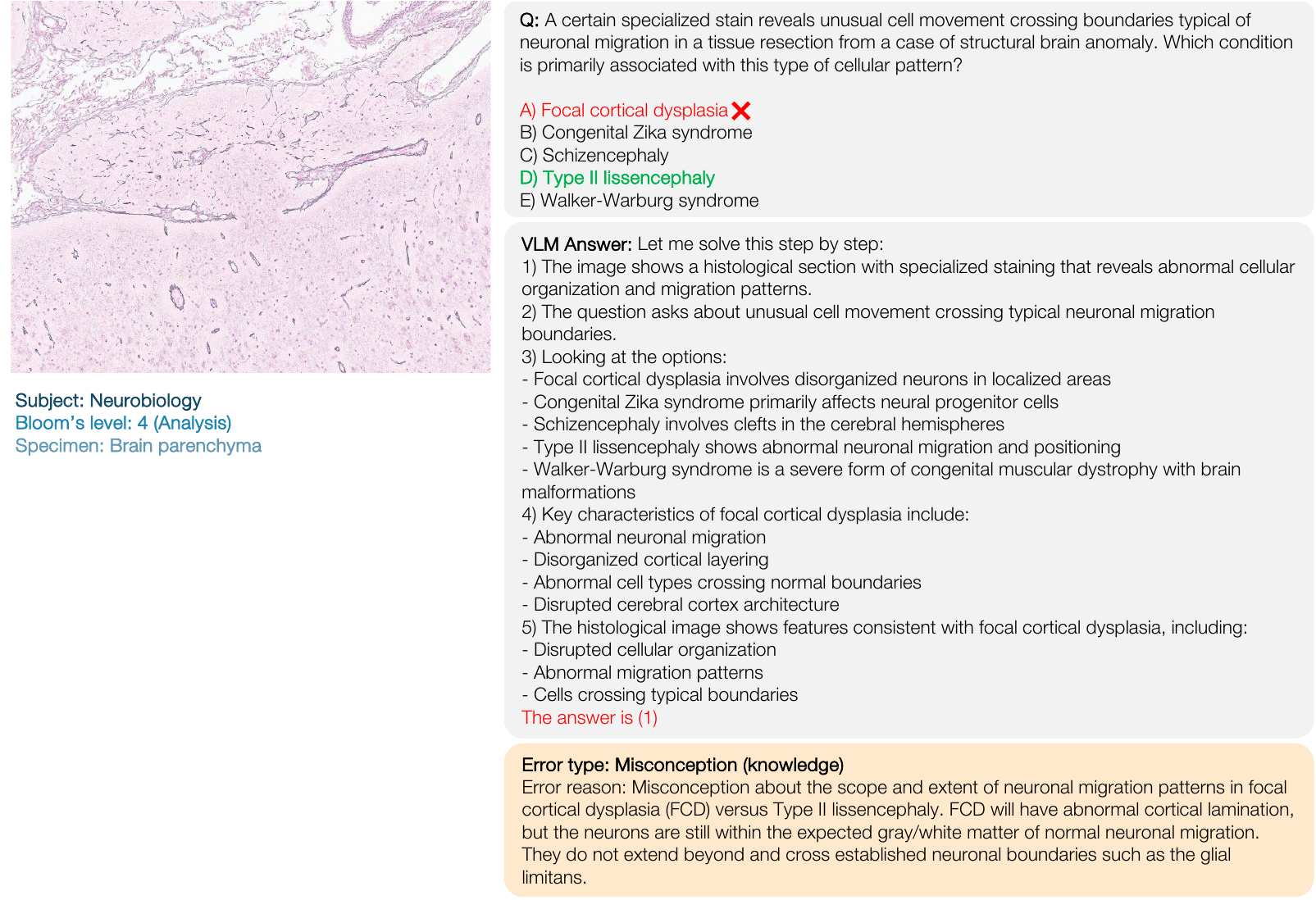}
    \caption{}
    \label{fig:error_knowledge_02}
\end{figure}

\newpage
\subsubsection{Overgeneralization or oversimplification errors}
\label{sec:appendix-error-analysis-overgeneralization}

\begin{figure}[h]
    \centering
    \includegraphics[width=\linewidth]{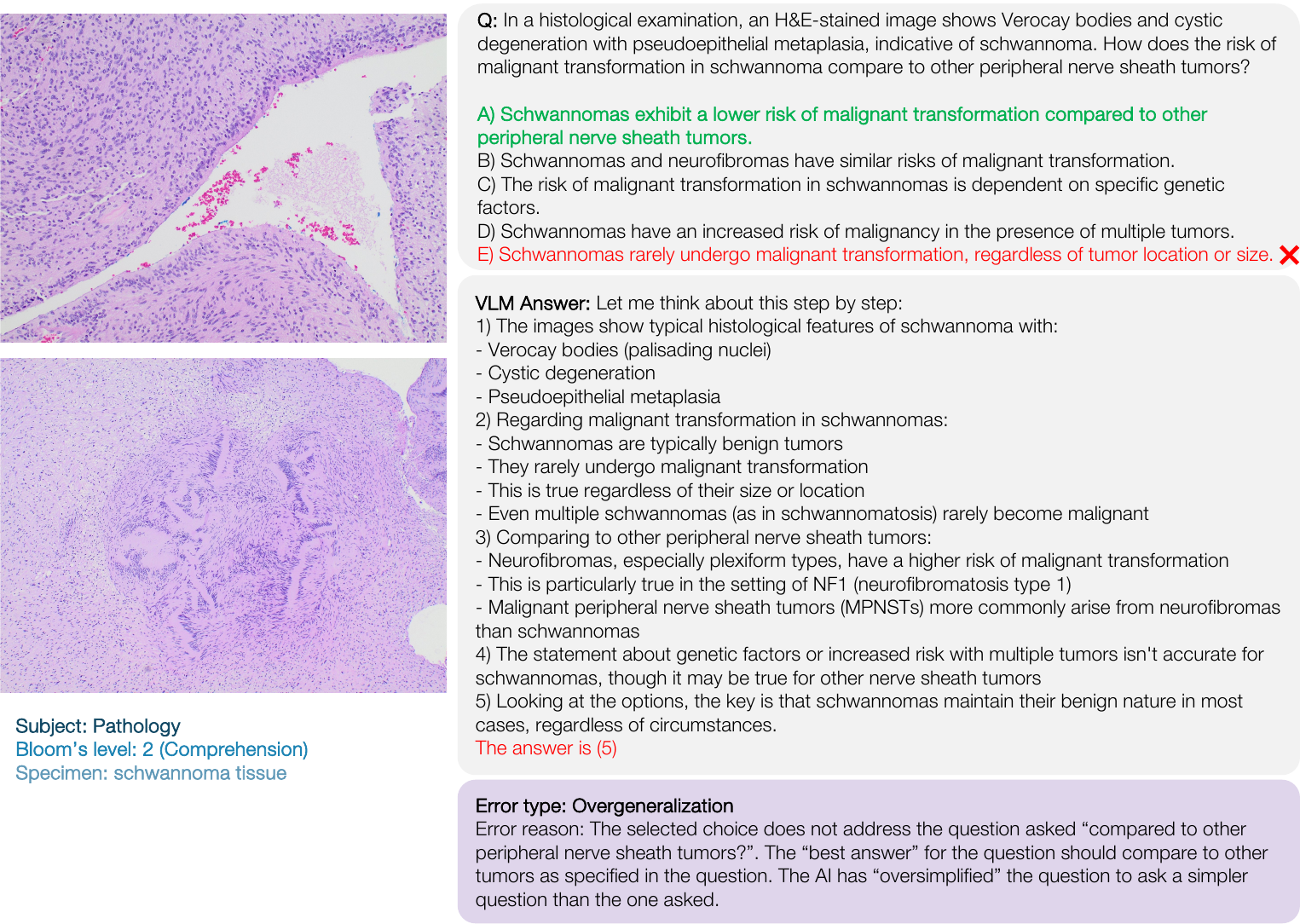}
    \caption{}
    \label{fig:error_generalization_00}
\end{figure}

\newpage
\begin{figure}[h]
    \centering
    \includegraphics[width=\linewidth]{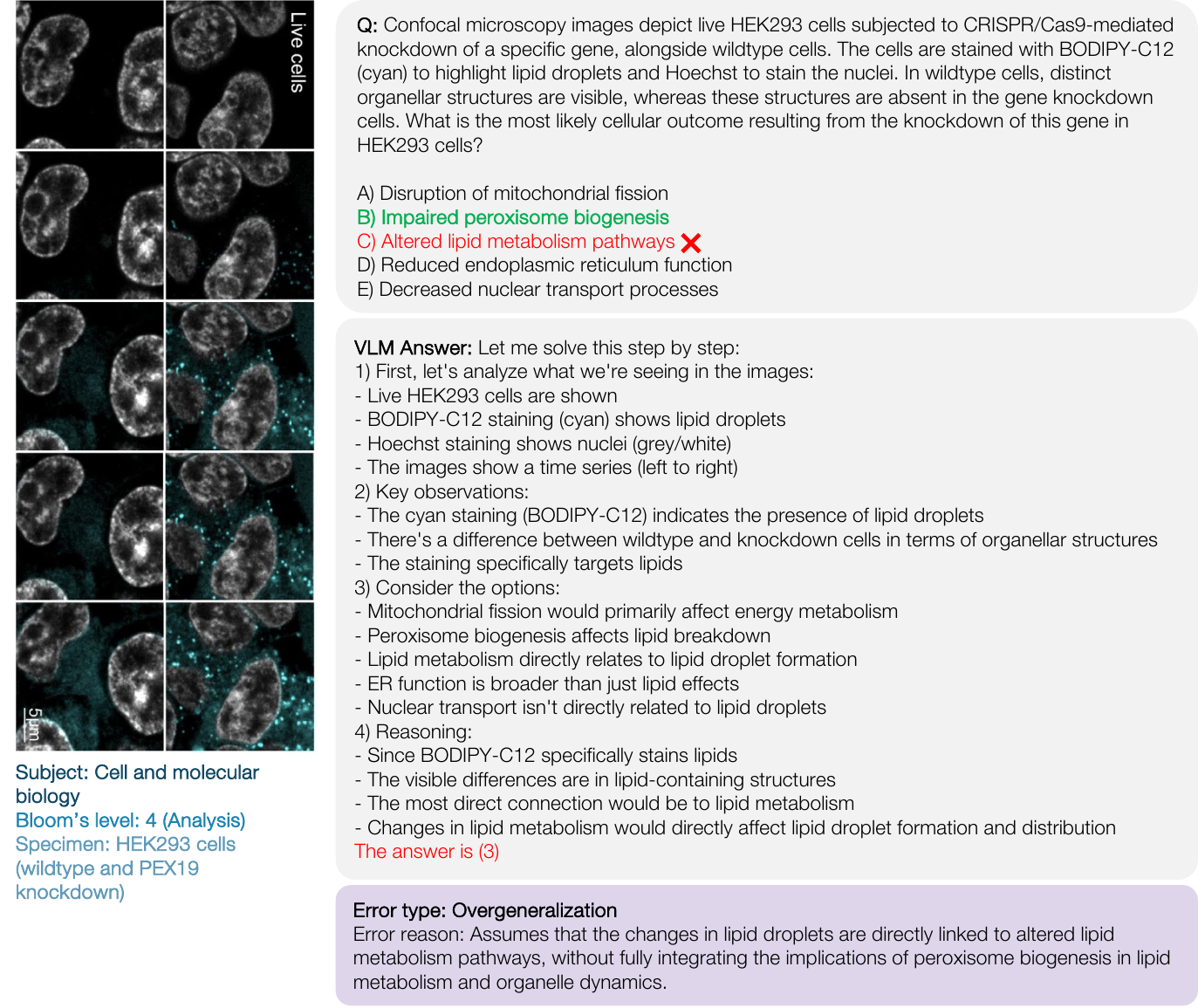}
    \caption{}
    \label{fig:error_generalization_01}
\end{figure}

\newpage
\subsubsection{Hallucination or errors due to added details}
\label{sec:appendix-error-analysis-hallucination}

\begin{figure}[h]
    \centering
    \includegraphics[width=\linewidth]{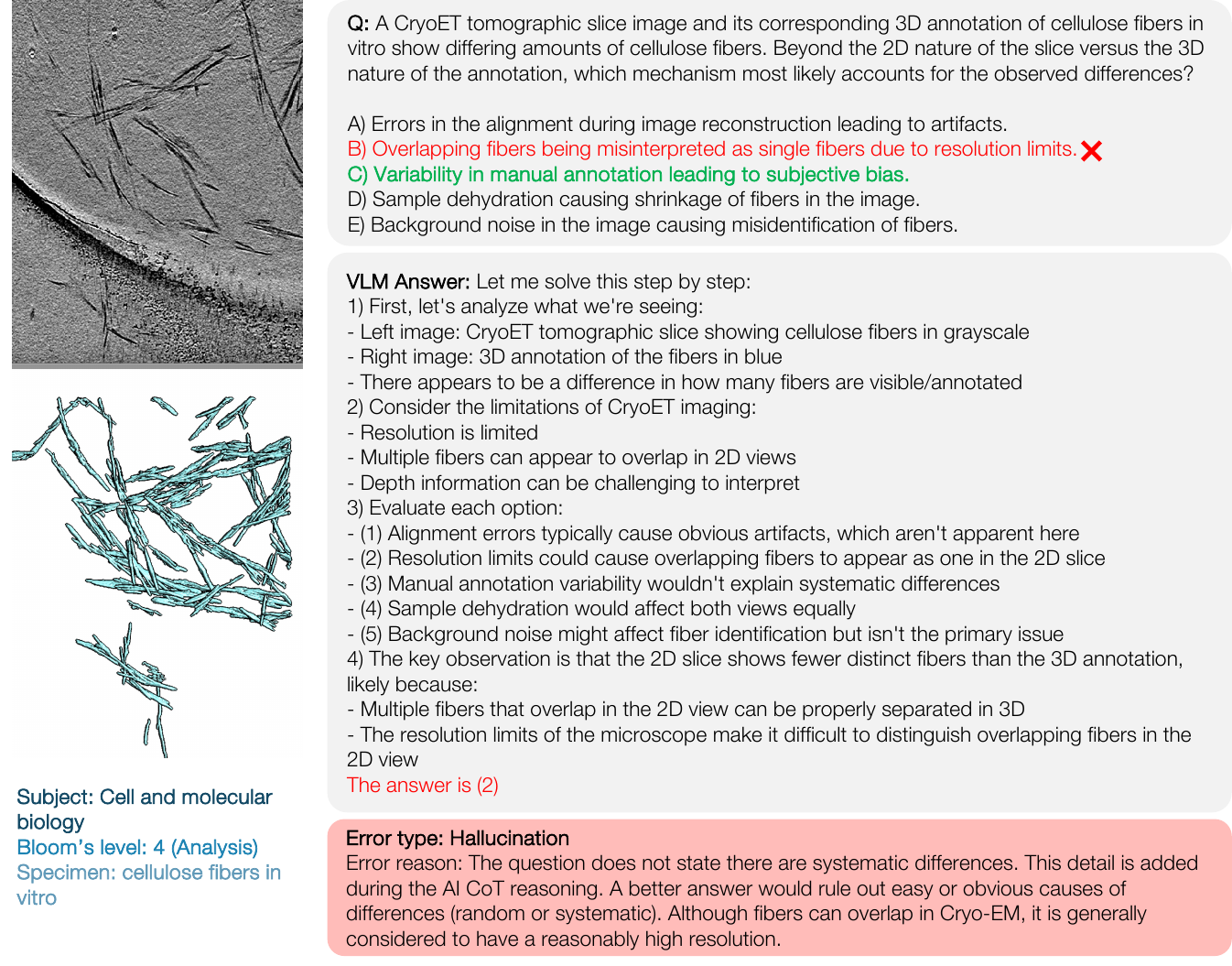}
    \caption{}
    \label{fig:error_hallucination_00}
\end{figure}

\newpage
\subsection{Reasoning trace analysis} 
\label{sec:appendix-reasoning-analysis}
\subsubsection{Human step-by-step reasoning} 
\label{sec:human-reasoning}
To better understand how subject matter experts approach questions in MicroVQA, we have a cell biologist and microscopist manually annotate their reasoning for each of the three main use cases. 

Reasoning example for task 1 `expert visual understanding'.
\begin{mdframed}[backgroundcolor=prompt_bg, font=\small, leftmargin=0cm, rightmargin=0cm]
\scriptsize
\begin{lstlisting}
*Question*
What subcellular structure could the bright green spots represent?*

*Answer*
This staining pattern most likely refers to cytoplasmic bodies.

*Reasoning:*
- Find green spots (no reasoning required)
- Rule out staining artefacts
    - Look at marker channels to assess whether cells appear intact
    - Compare the green spots to the marker channels for nuclei or microtubules
    - Check that green spots are always inside the cells
        - Yes, inside cells, so these are no staining artefact.
- Look up which spotty staining patterns in cells exist
    - Nuclear bodies
    - Nuclear speckles
    - Nucleoli
    - Cytoplasmic bodies
    - Centrosome
    - Vesicles
    - Intermediate filaments
    - Spotty microtubule staining
- Explore which patterns can be excluded
    - Check whether green spots are inside or outside of nuclei to know what patterns to explore
        - Outside of nuclei, thus nuclear subcellular localizations can be excluded, and cytoplasmic spotty patterns need to be explored
    - Check whether green spots are placed on microtubules
        - No, thus cannot be spotty microtubule stain
    - Assess localization > if accumulating around microtubule organizing center could be likely vesicles or centriolar satellites
        - No, not accumulating at centrosome
    - Assess size and shape of the spots
        - Small size and round, can exclude intermediate filaments
    - Count green spots per cell and check that not more than 10
        - Less likely vesicles, likely cytoplasmic bodies 
\end{lstlisting}
\end{mdframed}

Reasoning example for task 2 `hypothesis generation'.
\begin{mdframed}[backgroundcolor=prompt_bg, font=\small, leftmargin=0cm, rightmargin=0cm]
\scriptsize
\begin{lstlisting}
*Question*
The sperm cells swim in circles. Wouldn't they need to swim straight to reach the egg?

*Answer*
Sperm motility and its relation to reaching the egg and fertility is very complex and still not fully understood. Yet, it is known that sperm swimming in circles in a shallow observation chamber like you might have applied, may swim on helical or twisted-ribbon structures in an in vivo or 3D space and thereby might in fact efficiently swim forward.

*Reasoning*
- Explore the swimming trajectories to validate what the user means with swimming in circles.
    - Detect trajectory, assess how it would be described: curved or circular swimming trajectory
- Explore what is reported on swimming trajectories of sperm cells.
    - Sperm cell swimming trajectories are species dependent
        - Validate species: 
            - Assess whether the user mentioned the species in the prompt.
            - Assess look of cells in image (morphology, geometry) and compare to other images with known species eventually
                - Likely human sperm.
    - Sperm cell swimming trajectories are experimental setup dependent: shallow versus deep observation chamber
        - Validate setup: 
            - Brightfield microscopy, sperm cells are always swimming parallel, little to no off-focus blurring of the tail.
                - Assess how images from different experimental setups look like for human sperm
                    - Likely shallow cover-slip-based observation chamber
                        - Consequences of setup for trajectory
                            In shallow observation chambers sperm cannot freely beat and thus beat parallel to the glass surface, with no 3D component.
                            - Sperm cells hitting a glass surface align their beat with the glass and this way can get trapped if asymmetrically beating.
                            - Curved trajectory in 2D may relate to a progressive/straight swimming path in 3D
    - Sperm cell swimming trajectories are stimulus dependent and maturation-state dependent.
        - Validate setup and explore metadata provided: 
            - Check literature and compare image to available images of sperm from healthy donors (if available).
            - Compare detected swimming trajectory to swimming trajectories of sperm explored in a similar species and experimental setup where sperm cells are perturbed (if available).
            - Are there indications for any treatment under which the detected swimming trajectory would be abnormal and how would these look like (if available).
        - No indications for any treatment. Neither in prompt nor in image.
- Explore what is known on sperm cell trajectories to reach the egg in humans.
        - Not much is known since in vivo studies are not possible.
        - Sperm cells need to be able to develop hyperactive motility to swim to the egg
            - How do hyperactive sperm cells look?
                - Asymmetric beating
                    - What trajectory would they swim in in a 2D observation chamber?
                        - Curved trajectory
                            - Does that fit the swimming trajectory presented?
                                - Yes
- Assess whether when summarizing all the obtained information the conclusion from the observed swimming trajectory to sperm not reaching the egg is valid.
    - No.
- Compose summarizing reply to support the conclusion.
\end{lstlisting}
\end{mdframed}

Reasoning example for task 3 `experiment proposal'.
\begin{mdframed}[backgroundcolor=prompt_bg, font=\small, leftmargin=0cm, rightmargin=0cm]
\scriptsize
\begin{lstlisting}
*Question*
Can you suggest a simple experiment to find out whether centrosomal multiplication is occurring here?

*Answer*
One option could be to co-stain the EDC3 antibody with a centrosomal marker like PCNT or CEP250.

*Reasoning*
- Extract hypothesis to be tested from the question
    - EDC3 is labeling centrosomes and thus, centrosomal multiplication is occuring in the pictured cells
- Detect the experimental setup
    - Immunofluorescent antibody staining on a protein of interest (EDC3) in a human cultured cell type
- Compare to similar experimental setups that support such a hypothesis
    - Immunofluorescent antibody staining of centrosomes, co-labeling cell markers, performing microscopy, counting centrosomes
    - Acquiring or creating a transgenic cell line with a GFP-labeled centrosomal proteins, followed by microscopy
- Determine which methods are compatible with EDC3 staining
    - Both compatible with co-staining EDC3
- Determine which experiment is simplest for someone performing the shown experimental setup and if equal which one can be realized in less time
    - Immunofluorescent staining of EDC3 and centrosomes
- Find centrosome markers for human cells and assess suitability for this cell type
- Final decision on experimental setup
\end{lstlisting}
\end{mdframed}

\subsubsection{LLM-based error analysis} \label{sec:llm-error-analysis}
To better understand how the model approaches questions in MicroVQA, we extend our manual reasoning trace analysis from a subset of the dataset to all questions by leveraging zero-shot prompting of an LLM. The automatic error categorization achieves 63\% accuracy compared to manually assigned categories. Although imperfect, automated error categorization is scalable and preserves the overall distribution of human-identified errors, enabling broader analysis. We also classify correctly answered questions to quantify overall changes across experiments. Below, we present the prompts and category definitions used for classification.
\begin{mdframed}[backgroundcolor=prompt_bg, font=\small, leftmargin=0cm, rightmargin=0cm]
\scriptsize
\begin{lstlisting}
Below is a multiple choice question with options and the reasoning that lead a model to an incorrect response. Originally the model was also shown an image with the question. Your task is to use the reasoning trace to tag the error type:
- Perception: the image was not interpreted correctly.
- Overgeneralization: the details of the question were ignored and the general case was applied.
- Hallucination: details were added during reasoning that weren't in the question or extracted from the image.
- Other: the error does not fit the above categories.

Question:
{{question}}

Correct answer: {{correct_answer}}

Resoning trace:
{{reasoning}}
\end{lstlisting}
\end{mdframed}

\begin{mdframed}[backgroundcolor=prompt_bg, font=\small, leftmargin=0cm, rightmargin=0cm]
\scriptsize
\begin{lstlisting}
Below is a multiple choice question with options and the reasoning that lead a model to a correct response. Originally the model was also shown an image with the question. Your task is to use the reasoning trace and determine if the question was answered because of these reasons:
- No image: the image is unnecessary because the correct answer doesn't rely on interpreting visual cues.
- Visual giveaway: the image is unnecessary because critical visual information is already described in the question itself.
- Language bias: the question has information that makes the correct option obvious.
- Weak distractors: the distractors are easy to rule out according to the reasoning trace.
- Good question: the question is well-constructed and needs the image and specialized knowledge to answer.
- Other: the question is hard to answer or doesn't fit the other classes.

Question:
{{question}}

Correct answer: {{correct_answer}}

Resoning trace:
{{reasoning}}
\end{lstlisting}
\end{mdframed}
\begin{figure}[h]
    \centering
    \includegraphics[width=\linewidth]{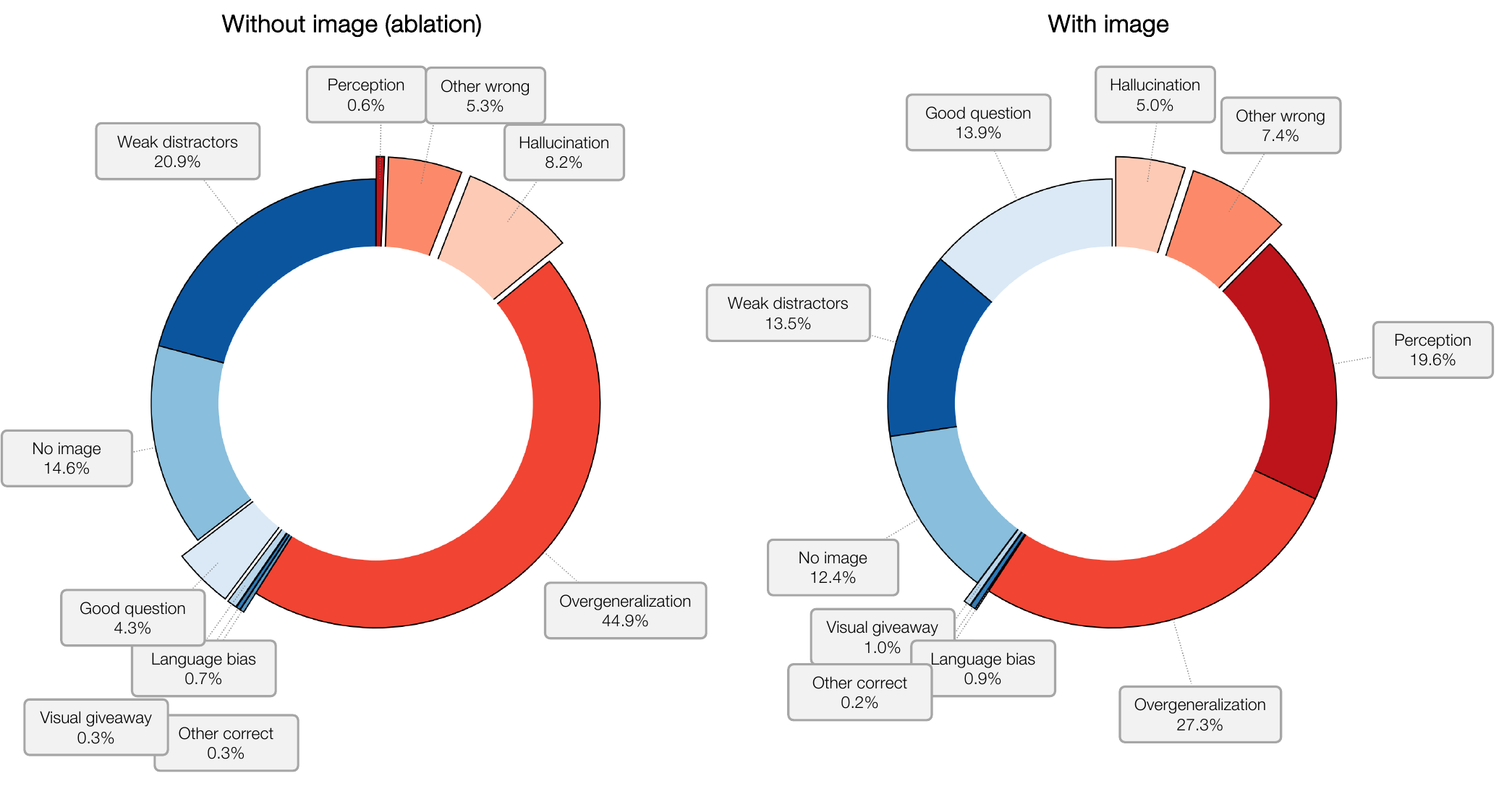}
    \caption{Reasoning trace categorization with an LLM (GPT-4o): Correct (blue) vs. incorrect (red) answers. Comparison of Claude-Sonnet’s responses without (left) and with (right) image access.}
    \label{fig:trace_analysis}
\end{figure}

We apply this system to examine the impact of providing or withholding the image for each question (Fig. \ref{fig:trace_analysis}). While overall accuracy remains comparable across both conditions, the distribution of errors and successful reasoning patterns differs meaningfully. As expected, hallucination errors decrease when the image is available, perception errors are rare when the model lacks visual input, and the proportion of questions that do not require an image remains stable.

Most errors arise from questions requiring visual and contextual understanding, highlighting the importance of multi-modal reasoning. Overgeneralization errors decrease when the image is provided, as the model is better grounded in the question context. However, many questions remain challenging even with visual input, as perception errors increase. This suggests that effectively solving our dataset requires models to both maintain the specifics of the question's context and accurately interpret visual information.

Additionally, note that the proportion of questions answered correctly through language shortcuts (e.g., language bias and visual giveaways) remains low in both settings, suggesting the effectiveness of the RefineBot strategy. However, addressing weak distractors remains a persistent challenge for specialized datasets.

\end{document}